\documentclass{article}

    \PassOptionsToPackage{numbers, compress}{natbib}

\usepackage[final, main]{neurips_2025}

\usepackage[utf8]{inputenc} 
\usepackage[T1]{fontenc}    
\usepackage{hyperref}       
\usepackage{url}            
\usepackage{booktabs}       
\usepackage{amsfonts}       
\usepackage{nicefrac}       
\usepackage{microtype}      
\usepackage{xcolor}         
\usepackage{graphicx}       
\usepackage{wrapfig}
\usepackage{float} 
\usepackage{amsmath}
\usepackage{array}
\usepackage{subfigure}
\usepackage{makecell}
\usepackage{siunitx}
\usepackage{comment}
\usepackage{multirow}

\title{Are Large Language Models Sensitive to \\ the Motives Behind Communication?}

\author{
Addison J. Wu$^{1*}$\quad
Ryan Liu$^{1*}$\quad
Kerem Oktar$^{2*}$\quad
Theodore R. Sumers$^{3}$\quad
Thomas L. Griffiths$^{1,2}$\\[2pt]
$^1$Department of Computer Science, Princeton University\\
$^2$Department of Psychology, Princeton University\\
$^3$Anthropic\\[4pt]
Code: \href{https://github.com/addisonwu05/LLM-Vigilance}{\texttt{https://github.com/addisonwu05/LLM-Vigilance}}
}

\begin{document}

\newcommand{\draftonly}[1]{#1}
\newcommand{\draftcomment}[3]{\draftonly{{\color{#2}{{{[#1: #3]}}}}}}

\newcommand{\ryan}[1]{\draftcomment{Ryan}{blue}{#1}}
\newcommand{\addison}[1]{\draftcomment{Addison}{red}{#1}}

\maketitle

\begin{abstract}
Human communication is \textit{motivated}: people speak, write, and create content with a particular communicative intent in mind. As a result, information that large language models (LLMs) and AI agents process is inherently framed by humans' intentions and incentives. People are adept at navigating such nuanced information: we routinely identify benevolent or self-serving motives in order to decide what statements to trust. For LLMs to be effective in the real world, they too must critically evaluate content by factoring in the motivations of the source---for instance, weighing the credibility of claims made in a sales pitch. In this paper, we undertake a comprehensive study of whether LLMs have this capacity for \textit{motivational vigilance}. We first employ controlled experiments from cognitive science to verify that LLMs' behavior is consistent with rational models of learning from motivated testimony, and find they successfully discount information from biased sources in a human-like manner. We then extend our evaluation to sponsored online adverts, a more naturalistic reflection of LLM agents' information ecosystems. In these settings, we find that LLMs' inferences do not track the rational models' predictions nearly as closely---partly due to additional information that distracts them from vigilance-relevant considerations. However, a simple steering intervention that boosts the salience of intentions and incentives substantially increases the correspondence between LLMs and the rational model. These results suggest that LLMs possess a basic sensitivity to the motivations of others, but generalizing to novel real-world settings will require further improvements to these models.

\end{abstract}

\begingroup
  \renewcommand\thefootnote{}   
  \footnotetext{\textsuperscript{*}Equal contribution. Correspondence to: Addison J. Wu \textless \texttt{addisonwu@princeton.edu}\textgreater, Ryan Liu \textless \texttt{ryanliu@princeton.edu}\textgreater, Kerem Oktar \textless \texttt{oktar.research@gmail.com}\textgreater.}
  \addtocounter{footnote}{-1}   
\endgroup

\section{Introduction}
\label{sec:1_introduction}

Much of the information available online---and hence a large fraction of the data large language models (LLMs) are tasked with processing---is the product of people's intentional communication: from op-eds in newspapers \citep{coppock_long-lasting_2018} to social media content by partisans \citep{cinelli_echo_2021} to word-of-mouth promotion \citep{kozinets_networked_2010} to even online reviews \citep{dixit2019integrated}. When navigating these digital environments, people naturally infer the accuracy of content. This capacity for tracking the processes that generate social data, known as epistemic vigilance \citep{sperber2010epistemic}, enables selective social learning. In particular, vigilance of others' motivations---\textit{motivational vigilance} \cite{oktar2025rational}---allows people to  track the intentions and incentives biasing data (for instance, people can ignore advice from malevolent sources while learning from benevolent ones). As LLMs are increasingly deployed in high-stakes environments---particularly as AI agents that act on behalf of users---it has become increasingly important to assess whether these systems are similarly vigilant to the motivations behind communication.

Recent research on LLMs suggests that they may have difficulty exercising such vigilance. Models are known to be vulnerable to jailbreaking, where ill-motivated instructions are followed and lead to manipulated outputs despite explicit guardrails~\citep[e.g.,][]{liu2024autodan, zeng2024johnny}. LLMs also demonstrate behavioral patterns such as sycophancy, where they express views that are aligned with a user's false beliefs rather than the truth~\citep{cotra2021alignment, perez2023discovering, sharma2024towards}. Separately, vision-language models and agents have been shown to be vulnerable to misleading stimuli in online environments such as pop-ups~\citep{chen2025evaluatingrobustnessmultimodalagents, zhang2024attacking} and distracting content~\citep{ma2024caution}. Underlying these behaviors are training paradigms which prioritize adherence to input instructions and user satisfaction, but not necessarily the vigilant monitoring of incentives and truth.

Yet, vigilance is key for LLM agents to act effectively on behalf of a user in real-world contexts \citep{stöckl2025aiagentsinteractingonline}. It enables LLMs to detect when information is generated by a motivated source, identify whether the source's motivations are benevolent vs. manipulative, and draw reasonable inferences given these social considerations. 
However, the current literature lacks a way to measure this ability in LLMs. 

\begin{figure}[htbp!]
    \centering
    \includegraphics[width=0.82\textwidth]{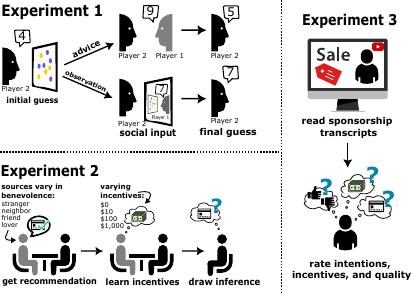}
    \label{fig:overview_figure}
     \caption{Our three experimental paradigms designed to assess different aspects of LLM vigilance: 1) Whether LLMs adjust their guess by \emph{discrminating} between directly motivated advice and incidentally observed social information. 2) Whether LLMs rationally \emph{calibrate} their vigilance to motivated communication by considering the speaker's benevolence and incentives. 3) Whether LLMs \emph{generalize} vigilance to realistic, context-laden YouTube sponsorship settings.}
\end{figure}

In our paper, we address this gap by leveraging established literature in social cognition to study whether LLMs exercise vigilance over motivated communication. We employ a cutting-edge rational model~\citep{oktar2024rational, oktar2025rational} from cognitive science as a \textit{normative benchmark} for motivational vigilance, and evaluate the capacity of LLMs to exercise vigilance across three experimental paradigms (Figure \ref{fig:overview_figure}). 

To assess whether LLMs can exercise motivational vigilance, we first examined if they are sensitive to the difference between motivated communication and incidental information --- a prerequisite for any vigilant behavior. We tested this using an adapted version of a two-player judgment task~\citep{WATSON2025106066}, where players solve different problems with the same answer. The player with the harder problem either receives a deliberate suggestion from the other player or secretly observes their answer. LLMs consistently shifted their answers closer when viewing the secretly observed answer, suggesting that \textbf{LLMs are capable of discriminating between deliberate and neutral information}, and appropriately modulate their belief updates given the motives behind communication. 

Next, we investigated if LLMs calibrate how much they update their beliefs by considering speaker incentives and intentions when receiving recommendations~\citep{oktar2024rational}. Specifically, we tested two paradigms: 1) when someone recommends a product to the user, who comes to the LLM for advice, and 2) when someone recommends a product to the LLM directly. Within each paradigm, we varied the trustworthiness of the person giving the recommendation and the known incentives that the person receives for recommending the product. We evaluated LLMs' inferences from these recommendations in controlled scenarios across finance, real estate, and medicine --- all domains in which generative AI systems are actively developed or deployed \citep[e.g.,][]{fitzpatrick2023generative,liu2024survey,nie2024survey}. Comparing LLMs' inferences to human data and rational models, we found that \textbf{non-reasoning LLMs draw substantially human-like} (Pearson's $r > 0.9$) \textbf{and approximately rational} ($r > 0.78$) \textbf{inferences in these simple structured settings}. However, reasoning models exhibit vigilance somewhat less reliably according to the rational model ($r \in [0.32, 0.72]$) and when compared to human data ($r \in [0.64, 0.87])$.

To examine whether their vigilance translates to ecologically valid inferences, we designed a new task in which LLMs draw inferences about the quality of products from 300 randomly sampled sponsored advertisements in YouTube videos, from NordVPN to AG1 nutrition. \textbf{LLMs were much worse at drawing vigilant inferences in these naturalistic contexts} ($r < 0.2$). Further experiments revealed that this drop in performance is partly a consequence of LLMs failing to ground their inferences in the trustworthiness and incentives of the speaker when given noisy inputs. However, vigilance-based prompt steering shows promise as an avenue for improving model performance. 
Together, our results show that LLMs possess a basic sensitivity to the motivations of others, but generalizing this sensitivity to novel, real-world settings will require additional improvements to these models. 

\section{Related Work}
Our work draws on three distinct strands of research: first, studies in social cognition focusing on the mechanisms and properties of motivational vigilance in humans; second, research on LLMs that measures related social capabilities; and finally, a broader line of work that uses cognitive science to evaluate and understand LLMs. We cover each in turn.

\subsection{Motivational vigilance in humans}
\label{sec:2_1_vigilance}
People are naturally sensitive to the reliability of others as information sources \citep{mascaro_moral_2009,oktar_how_2025}. This sensitivity is a pre-requisite for effective social learning: given a mixture of reliable and deceptive sources, neither blind trust nor complete dismissal supports adaptive inference \citep{mercier_not_2020,mascaro_moral_2009}. Such vigilance is essential for downstream behaviors, such as disagreement resolution \citep{oktar2024dimensions,wagner-pacifici_resolution_2012} or detecting deception \citep{bond_jr_overlooking_2013,vrij_detecting_2000}. Research on social cognition has identified this capacity as comprised of two components: vigilance of \emph{competence} (whether the source is knowledgeable) and vigilance of \emph{motivations} (whether the source is acting benevolently) \citep{sperber2010epistemic}. 

Research in psychology has primarily focused on the former---how percieved competence influences social learning \citep{harris_trusting_2012,jaswal_adults_2006,koenig_characterizing_2011,sobel_knowledge_2013}. However, emerging work on the mechanisms underlying motivational vigilance has identified two key factors in people's judgments: a speaker's \emph{intentions} (whether they seek to altruistically benefit the listener or selfishly manipulate them) \citep{landrum_learning_2015}, and \emph{incentives} (whether the speaker stands to benefit from being deceptive) \citep{bond_jr_overlooking_2013}. Attending to these factors can mitigate malevolent manipulation (e.g., scams, Ponzi schemes, lies, and deceit), although succesful manipulators can in turn circumvent vigilance by appealing to reciprocity or engaging in relationship-building \citep{cialdini_social_2004}. This demonstrates how strategic communication \citep{maschler2020game} relies on recursive social inference: Listeners reason about why speakers are choosing particular utterances, and speakers choose utterances based on what listeners are likely to infer \citep{kamenica2011bayesian,okeefe_relative_2013,wood_attitude_2000}. Such inferences require  an understanding of others' minds and subjective representations of the world, known as \textit{Theory of Mind} \citep{baron1985does,frith2005theory,leslie_core_2004}.

\subsection{LLM failures and inferring communicative intent}
\label{sec:2_2_LLM_failures}

Recent LLMs have been aligned using Reinforcement Learning from Human Feedback \citep{ouyang2022training, touvron2023llama}, which leverages human data to generate more desirable responses that align with human preferences. However, this training can also introduce various undesirable effects such as increased hallucinations~\citep{ouyang2022training}, reward hacking~\citep[e.g.,][]{denison2024sycophancy,hendrycks2021unsolved, singhal2024a}, and deception~\citep[e.g.,][]{liang2025rlhs,williams2025on}. 

A series of failure effects previously observed in LLMs can be framed as a lack of motivational vigilance. Models are vulnerable to jailbreaking~\citep{liu2024autodan,zeng2024johnny}, where an ill-motivated user's instructions are followed --- leading to undesirable outputs. One type of sycophancy is the tendency for LLM responses to follow user beliefs over the truth~\citep{cotra2021alignment, perez2023discovering,sharma2024towards}, which can occur because LLMs lack a deeper understanding of user intent.
In both cases, exercising more vigilance over the belief state and motivations of the user could help mitigate harmful outputs. More immediately, prior work has found that ill-intentioned information in online environments such as pop-up windows~\citep{zhang2024attacking} and distractions~\citep{chen2025evaluatingrobustnessmultimodalagents,ma2024caution} harm the ability for multimodal models to complete agentic tasks. Underlying these behaviors are training paradigms which prioritize adherence to user preferences in local interactions, without accounting for the complex and nuanced aspects of real-world strategic communication.

Vigilance is also linked with other evaluated capacities of LLMs: It can be viewed as an input to social behaviors such as conformity~\citep{asch1951conformity, weng2025we, zhu2024conformity}, where one's vigilance to others’ ill-intent can be used to decide whether to conform. Other social capacities are precursors to vigilance: LLMs can exhibit false beliefs~\citep{chen2024tombench,kosinski,shapira2024clever} that a speaker is malicious, leading to incorrect priors for how much they should be trusted. LLMs could also misinterpret the communicative intent of an utterance~\citep{shapira2023well, yi2025irony, zhao2026improving}, leading to faulty inferences about the utterance itself. However, vigilance stands apart from these capacities, connecting one's priors about a speaker's trustworthiness and incentives to how much one should update their beliefs from the speaker's words.

\subsection{Using cognitive science to study LLMs}
\label{sec:2_3_LLM_cognitive_science}
A broad line of work applies cognitive science to help understand LLMs~\citep{ku2025using}, typically leveraging controlled tasks and carefully curated stimuli to test specific behavioral hypotheses. The compatibility of study stimuli and availability of human data allow these evaluations to transfer to LLMs with minimal changes \citep[e.g.,][]{binz2023using, pmlr-v235-coda-forno24a}. 
In recent years, this interdisciplinary line of work has studied various aspects of LLMs, including their representational capacity and alignment \citep{frank2023baby, peterson2018evaluating}, inference time reasoning \citep{liu2025mind,prystawski2022psychologically}, social biases \citep{doi:10.1073/pnas.2416228122}, episodic memory \citep{cornell2023role}, and theory of mind \citep[e.g.,][]{kosinski, sap-etal-2022-neural, shapira2024clever,ullman2023large}.

A subset of this literature focuses on intersections between LLMs and psychological theories of rationality. Past work has used rational models of decision-making to study LLMs' probability judgments \citep{zhu2024incoherent} and assumptions of human behavior \citep{liu2025large}. Resource rationality, describing the trade-off between expected utility and computation cost~\citep{kahneman2011thinking,lieder2017strategy}, has also been used to understand and guide LLM outputs~\citep{ cherep2024superficial,de2024rational}. Such rational communicative models have also been used to study value conflicts in LLMs \citep{liu2024largea}, and a similar line of work studies LLMs' economic rationality using hypothetical scenarios and economic games \citep{chen2023emergence, huang2025competing, raman2024steer, ross2024llm}. Our work follows this general approach, using rational models from cognitive science to examine LLMs' vigilance to motivated communication.

\section{Experiment 1: Can LLMs discriminate between deliberately communicated and incidentally observed information?}
\label{sec:exp_1}

Researchers distinguish between two major categories of social information with contrasting motives \citep{HENRICH2009244, valone2007public, WATSON2025106066}. The first is \emph{deliberately communicated} information, intended to influence a listener, such as an online promotion for a company's stock. The second is \emph{incidentally observed} information, revealed without a direct intent to persuade, such as a diary entry. Contrasting these forms of information is the simplest test of vigilance: whether LLMs can use the generative function behind the data---deliberate communication or first-order evidence \citep{hedden_almost_2022}---to modulate inference.

\subsection{Experimental setup} To evaluate whether LLMs are sensitive to this difference, we adapt an experimental paradigm from Watson and Morgan \cite{WATSON2025106066}. Each trial presents two players with images of mixed blue and yellow circles (see Figure~\ref{fig:overview_figure}), and their task is to compute the difference between the number of blue and yellow circles within 2 seconds. Player 1 completes the task first for 20 easy images, then Player 2 completes the task for 20 hard images that share the exact same answer list (see Figure~\ref{fig:exp1_exemplar_diagrams}).

When completing the task, Player 1 is randomly assigned to give either ``advice'' in the form of a number (deliberate communication) or their ``spied'' actual answer (incidentally revealed) to Player 2 for each image. No other communication is allowed. After Player 2 provides their initial answer, they are shown this number and if it was ``advice'' or ``spied'', and are allowed to revise their guess. We test this across payoff structures ranging from cooperative (both players are rewarded if at least one player is correct) to competitive (only the player with the most correct answers receives a reward). 

\subsection{Models, hyperparameters, and prompts} We conduct evaluations on GPT-4o and Claude 3.5 Sonnet. For each model, payoff structure, and prompting method, we conduct $n = 30$ trials over the same 20 pairs of images (order shuffled every trial), with temperature $= 1$. 
In all trials, the same LLM was assigned to both Player 1 and 2. 

While the original experiment used a time constraint to induce uncertainty in human answers, this does not apply to models. This constraint was important because if Player 2 was confident about their answer, they would not be influenced by any information provided by Player 1. To induce uncertainty in models, we instead made the task more difficult by adding noise to the images (Appendix \ref{app:exp1_images}) and prompting models' initial guesses directly. We confirm that this setup indeed reduces model accuracy (see Section~\ref{sec:3_3_results}), allowing us to measure vigilance effectively. Moreover, this does not affect our study of vigilance, as it can be exercised optimally or suboptimally independent of the properties of the reasoner (e.g., time, compute, memory).
Player 1's advice and Player 2’s response to the advice/spied information from Player 1 were generated both directly and with chain-of-thought (CoT) \citep{wei2022chain}. We limited output tokens to 10 for direct and 750 for CoT (prompts in Appendix \ref{app:exp1_prompts}).

\subsection{Results} 
\label{sec:3_3_results}
\begin{figure}
    \centering
    \begin{minipage}{0.7\textwidth}
    \includegraphics[width=\linewidth]{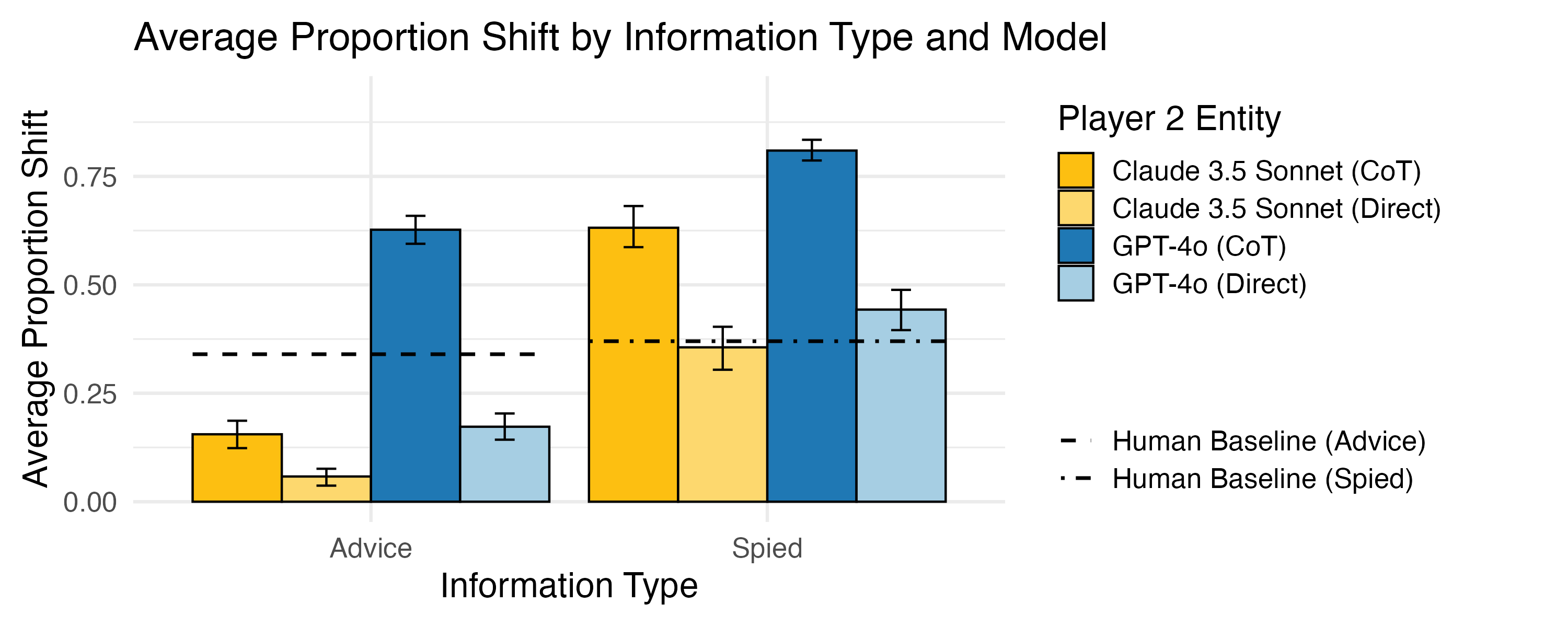}
    \end{minipage}
    \begin{minipage}{0.29\textwidth}
    \caption{On average, LLMs as Player 2 shifted their initial estimates more when viewing incidental (spied) information compared to deliberate advice. CoT increased such shifts in all conditions.
    }
    \label{fig:exp_1avg_prop_shift}
    \end{minipage}
\end{figure}

\paragraph{LLMs can
discriminate between deliberate vs.~incidentally observed social information.} As shown by the average proportion shifts in Figure \ref{fig:exp_1avg_prop_shift}, when playing as Player 2, both GPT and Claude statistically significantly changed their score less when receiving deliberate advice from Player 1 than if they had ``spied'' Player 1's true answer. This was consistent with human participants~\citep{WATSON2025106066}. Like humans, LLMs also took into account the incentives of Player 1, adjusting their score more when rewards were cooperative rather than competitive---at an even higher sensitivity than people (Appendix \ref{app:exp1}, Figure \ref{fig:exp1_social_shift_payoff}). These results demonstrate that LLMs can identify motivated communication and react according to others' incentives, thus satisfying a basic capacity for motivational vigilance. 

\paragraph{Experimental design successfully elicits vigilant behavior.} Separately, we confirm that our replacement constraint allows for vigilance in Player 2's response: LLMs with the adjusted constraint had first-guess accuracy rates from 18--44\%, slightly lower than human participants in the original setting (55\%). Being correct did not significantly affect how much LLMs updated their answers (Appendix~\ref{app:exp1_stats}, Table~\ref{table:trust_pvalues}), implying that this numerical difference is not of concern. Thus, our design facilitates similar vigilance reasoning in LLMs as the original design does for human participants.

\paragraph{CoT prompting makes LLMs more likely to change their answers.} Interestingly, CoT encourages LLMs to exhibit greater susceptibility to Player 1's input (Figure~\ref{fig:exp_1avg_prop_shift}; full distributions in Appendix \ref{app:exp1_prop_shift}), leading to significantly larger shifts in Player 2's estimates towards Player 1 ($>60\%$ in 3 of 4 cases). These higher influence magnitudes deviated greatly from human participants (0.34 for advice, 0.37 for spied). This suggests that while CoT may enhance reasoning, it can also inadvertently amplify trust in social information, yielding patterns that are less aligned with motivational vigilance.

\section{Experiment 2: Can LLMs exercise nuanced vigilance towards motivated communication?}
\label{sec:exp_2}

Our previous experiment demonstrated that LLMs draw different inferences from incidental observation and motivated communication. But can these models account for the complex qualitative and quantitative dimensions of strategic communication when deciding who---and how much---to trust others? Answering this question requires formalizing a normative benchmark for rational inductive inference from motivated communication. For this, we employ the most advanced rational model from psychological literature: Oktar et al.~\citep{oktar2024rational, oktar2025rational}, which we use as a standard for motivational vigilance. This model extends models of pragmatic \citep{goodman_pragmatic_2016} and instrumental communication \citep{sumers_reconciling_2023} by considering informants with diverse intentions and incentives---improving upon past research which primarily focused on inference from purely helpful \citep{degen_rational_2023} or deceptive \citep{oey_designing_2022} sources. Here, we first describe the model, and then use it to measure LLM vigilance.

\subsection{Rational Model} Oktar et al. \citep{oktar2024rational, oktar2025rational} formalize vigilance as a form of recursive social inference, where listeners reason about speakers' intentions and incentives to identify whether their motivations are truth-promoting or manipulative. The speaker, in turn, reasons about the listener to identify which utterances are most likely to achieve their communicative goals. 
The speaker's probability of choosing an utterance $u$, $P_S(u)$, depends on the utility of the utterance, which is given by the `joint' reward $R_\text{Joint}$ associated with each subsequent listener action $\alpha$. $R_\text{Joint}$ is modeled as a combination of the speaker's outcomes, $R_S$, and the listener's outcomes, $R_L$, weighed by the speaker's benevolence, $\lambda$:
\begin{equation}
\label{joint_reward_eq}
    R_\text{Joint}(R_L, R_S, \lambda, a) = \lambda R_L(a) + (1-\lambda)R_S(a),
\end{equation}
where $\lambda \in [0, 1]$. When $\lambda=0$ the speaker is purely self-interested, and considers only their personal instrumental reward. When $\lambda=1$ the speaker is purely altruistic.

The speaker combines this joint reward with the probability that a non-vigilant, purely literal listener would follow an action given the utterance---which is given by the listener's policy, $\pi_L(a)$:
\begin{equation}
\hspace{-2mm}P_S(u \mid R_S, R_L, \lambda, A) \propto \exp \{\beta_S \cdot\sum_{a\in A} R_\text{Joint}(R_L, R_S, \lambda, a) \pi_L(a \mid u)\}.
\end{equation}
The vigilant listener uses this process to identify the probability that a recommended option in fact carries the advised reward, $P_{L}(R_L \mid u)$, by using their prior information to marginalize out other considerations (i.e., $P(R_s),P(R_L),P(\lambda)$):
\begin{equation}
P_{L}(R_L \mid u) \propto P_S(u \mid R_S, R_L, \lambda, A)P(R_s)P(R_L)P(\lambda).
\label{eq-vigilance}
\end{equation}
Further details of the model are in \citep{oktar2024rational, oktar2025rational}. For our purposes, this rational model provides quantitative predictions that we can use as a normative benchmark to rigorously evaluate LLM vigilance.


\subsection{Experimental setup} To test whether LLMs exhibit vigilance to this more refined degree, we turn to a second psychological paradigm from Oktar et al.~\citep{oktar2024rational, oktar2025rational}. In this experiment, all information is deliberately communicated, but the speaker's incentives and benevolence (manipulated via social closeness to the listener) vary systematically. 
We examine three decision settings in which participants are recommended an option individually by four characters, with each character receiving one of four possible incentive values, known to the LLM listener. All scenarios, characters, and incentives are described in Appendix \ref{app:exp2_details}.

Following the structure of human experiments \citep{oktar2024rational, oktar2025rational}, in each setting, each of the 16 different character-incentive speakers provides a recommendation for a particular option, and we subsequently elicit LLMs' opinions on the quality of that option. This allows us to measure `influence scores,' which capture the extent to which the recommendation of each speaker influences LLMs' beliefs about the reward of that option (i.e., $P(R_L|u)$). For calibration, scores for all 16 speakers are elicited within the same multi-turn dialogue, with all elicitations after the first speaker phrased counterfactually. The order in which speakers are presented is randomized. We used the same procedure to elicit `incentive scores,' which track the perceived goodness of different incentives for speakers to say certain utterances ($R_S$), and `trust scores,' which capture the perceived benevolence of the four characters (i.e., $\lambda$). We prompted for each score type in independent context windows.

We introduce additional conditions which vary prompts across reasoning elicitation --- direct vs.~CoT --- and roles --- LLM as the listener itself (first-person/agent perspective) vs.~LLM aiding a listener/user (assistant perspective). These allow us to examine sensitivity to motivations under a broader set of contexts in which LLMs are commonly used. A full list of prompts is in Appendix \ref{app:exp2_details}.

\subsection{Models and hyperparameters} We evaluate a suite of state-of-the-art LLMs: GPT-4o, Claude 3.5 Sonnet, Gemini 2.0 Flash, and Llama 3.3-70B, smaller open models: Llama 3.1-8B, Llama 3.2-3B, and Gemma 3-4B, and reasoning models: o1, o3-mini, and DeepSeek-R1. Models can output up to 10 tokens for direct prompts and 750 tokens for CoT or reasoning models. All models are sampled at temperature $1$. We prompt most models $n=40$ times for each setting and prompt condition. Reasoning models were instead prompted $10$ times due to cost, while GPT-4o was prompted $80$ times due to available Azure credits.

\subsection{Results} 
\begin{table}[h!]
 \caption{Average correlations across all prompting conditions for each model.}
 \label{table:exp_2_corrs1}
 \centering
 \resizebox{0.75\textwidth}{!}{
 \begin{tabular}{lc|cc}
   \toprule
   {} & \multicolumn{3}{c}{Correlation} \\
   \cmidrule(r){2-4}
   Model & Bayesian–LLM & Bayesian–Human & LLM–Human \\
   \midrule
   GPT-4o & $0.911$ & $0.929$ & $\mathbf{0.943}$ \\
   Claude 3.5 Sonnet & $0.845$ & $0.889$ & $\mathbf{0.941}$ \\
   Gemini 2.0 Flash & $0.788$ & $0.901$ & $\mathbf{0.925}$ \\
   Llama 3.3-70B & $0.876$ & $\mathbf{0.923}$ & $0.922$ \\
   \midrule
   o1 & $0.705$ & $\mathbf{0.894}$ & $0.861$\\
   o3-mini & $0.716$ & $\mathbf{0.869}$ & $0.712$\\
   DeepSeek-R1 & $0.326$ & $0.492$ & $\mathbf{0.643}$\\
   \midrule
   \midrule
   Llama 3.1-8B & $0.608$ & $\mathbf{0.813}$ & $0.701$ \\
   Llama 3.2-3B & $0.349$ & $\mathbf{0.586}$ & $0.550$ \\
   Gemma 3-4B & $0.288$ & $\mathbf{0.340}$ & $0.266$ \\
   \bottomrule
 \end{tabular}
 }
\end{table}

\paragraph{Frontier LLMs exercise consistent vigilance in controlled scenarios.} To measure vigilance, we examine whether LLMs' trust scores (i.e., $\lambda$) modulate their influence scores (i.e., $P(R_L|u)$) in accordance with the Bayesian rational model. Specifically, we measure the Pearson correlation between the LLM's elicited influence scores and the value generated by the rational model when fitted to the LLM's elicited trust and incentive scores. For brevity, we report averaged values over settings and prompt conditions, and provide full results in see Appendix~\ref{app:exp2_full_results}. 

We find that frontier non-reasoning LLMs (GPT-4o, Claude 3.5 Sonnet, Gemini 2.0 Flash, Llama 3.3-70B) exercise internally-consistent motivational vigilance, with correlations to the Bayesian model around $0.8$ to $0.9$ (Table \ref{table:exp_2_corrs1}, left column). Of these, GPT-4o demonstrates the highest levels of internal rationality ($0.911$). For these LLMs, we observe consistently high correlations across prompts---CoT vs. Direct---and roles---as a principal assistant, or the participant itself (see Appendix~\ref{app:exp2_full_results}). 

\paragraph{Frontier non-reasoning LLMs combine their priors to be more humanlike than rational models.} We also observe that at a statistically significant level ($p < .05$), humans' influence scores correlate \textit{better} with these LLMs' influence scores than the rational model fit on these same LLMs' elicited priors (right vs.~middle column, Table~\ref{table:exp_2_corrs1}). 
This suggests that these LLMs may capture residual variance in human vigilance beyond that which can be explained through a rational analysis, such as heuristics or biases that people employ when evaluating advice.
For frontier non-reasoning models, this finding was consistent in over 90\% of individual prompt conditions and roles (see Table~\ref{app:exp2_full_results}), but was only half as often the case for reasoning models, and never the case for smaller models.

\paragraph{Reasoning models are less vigilant.} Comparatively, reasoning models are less correlated with the rational model fitted on their priors (Table~\ref{table:exp_2_corrs1}, left column). O-series models' correlations are around $0.7$, while DeepSeek-R1 had a much lower average correlation around 0.3. Notably, these models were much less vigilant in the user assistant role than first-person; o-series models saw a drop in correlation around $-0.1$, while DeepSeek's correlation reduced from $0.793$ to $-0.141$ (see Appendix~\ref{app:exp2_full_results}), representing complete insensitivity to character trustworthiness and incentives.

We also perform analyses isolating the effects of different factors (i.e., characters vs.~incentives) on vigilance (Appendix~\ref{app:exp2_analyses}). We found that all three reasoning models tested performed among the worst in correlations along both dimensions. This suggests that reasoning models may currently be ill-prepared to take on agentic tasks in environments containing ill-motivated communication.

\paragraph{Vigilance improves with LLM scale and capability.} We find that model scale has a significant impact on the capacity for LLMs to exercise vigilance. Smaller LLMs’ (Llama 3.2-3B, Llama 3.1-8B, Gemma 3-4B) judgments were much less correlated with the rational model than frontier LLMs (around 0.3 to 0.6; Table~\ref{table:exp_2_corrs1} left column). While frontier models differed from the Bayesian model in more humanlike ways, smaller LLMs also correlated worse with human behavior (Table~\ref{table:exp_2_corrs1}, right column), suggesting that the low correlations observed are a result of a direct lack of vigilance capabilities---in being able to consolidate relevant priors into a coherent, vigilant analysis of advice.



\section{Experiment 3: Do LLMs generalize vigilance to naturalistic online settings?}
\label{sec:exp_3}

Experiments in cognitive science and psychology are simple and abstract so as to reliably identify the effect of specific manipulations on judgments or reasoning. Although the experiment of Oktar et al. \citep{oktar2024rational, oktar2025rational} is grounded in a socially meaningful setting, its delivery remains highly controlled and vignette-based, limiting ecological validity. Psychological behaviors observed in controlled settings can break down in the real world \citep{holleman2020real, johnson1972reasoning, yarkoni2022generalizability}, and given that LLMs will ultimately be deployed in open-ended interactive environments \citep{Bommasani2021FoundationModels, liu2024agentbench, zhou2024webarena}, we must ask whether LLMs are able to transfer motivational vigilance to a more ecologically valid domain.

\subsection{Dataset creation} To construct a set of ecologically valid stimuli, we consider real online product sponsorships and promotions, which naturally reflect financially motivated communication. We first obtain a comprehensive dataset of existing sponsorships on the video-hosting website YouTube from SponsorBlock \citep{sponsorblock}. We then use the YouTube Data API \citep{google_youtube_api} to scrape data for 300 randomly selected video IDs, containing the video title, channel name and description, and the transcript text extracted from the corresponding timestamps in the SponsorBlock dataset. To mitigate the confounding effects caused by an LLM's existing impressions of a specific product, we censor out all explicit mentions of brand and product names from transcript excerpts using GPT-4o (see Appendix \ref{app:exp3_prompts} for details). 

\subsection{Experimental setup} Analogous to the approach in Section \ref{sec:exp_2}, for each video sponsorship segment, we prompt an LLM to elicit its beliefs about the quality of the product promoted in the sponsorship ($P(R_L|u)$), how beneficial it perceives the sponsorship deal was for the corresponding YouTube channel ($R_S$), and how trustworthy it perceives the YouTube channel with respect to the viewer's wellbeing ($\lambda$). Queries for each video sponsorship were conducted in independent context windows, and for each sponsorship, each of the three variables ($P(R_L|u), R_S, \lambda$) were also prompted in separate context windows. We again examine the effects of prompting for reasoning elicitation --- direct vs. CoT --- and perspective --- LLM as the listener/agent itself (first-person perspective) vs.~LLM aiding a listener/user (user perspective). Specific prompts can be found in Appendix \ref{app:exp3_prompts}. 

\subsection{Models and hyperparameters} To obtain broader coverage over a larger number of sponsorships, we examined only GPT-4o, Claude 3.5 Sonnet, and Llama 3.3-70B and queried each video and prompting combination $n = 1$ time with temperature $= 0$ to minimize variability. We allowed each model to output at most 10 tokens for direct outputs, and 750 tokens for CoT outputs.

\begin{table}[h!]
  \caption{In realistic settings, LLMs' correlations to the Bayesian model greatly decrease. However, a prompt steer targeting speaker incentives recovers some rationality. Higher values in each row are in \textbf{bold}, and * denotes statistically significant improvement ($\alpha=.05$) compared to the default prompt.}
  \label{table:youtube_steering}
  \centering
  \resizebox{0.86\textwidth}{!}{
  \begin{tabular}{lSS}
    \toprule
    {} & \multicolumn{2}{c}{Correlation with Bayesian inference model} \\
    \cmidrule(r){2-3}
    LLM/Prompting Combination & {Default Prompt} & {Steering Prompt} \\
    \midrule
    GPT-4o CoT First-Person         &  0.024  &  $\mathbf{0.137}$* \\
    GPT-4o CoT User                 &  0.008  &  $\mathbf{0.143}$* \\
    GPT-4o Direct First-Person     &  0.121  &  $\mathbf{0.234}$* \\
    GPT-4o Direct User             & -0.006  &  $\mathbf{0.312}$* \\
    \midrule
    Claude 3.5 Sonnet CoT First-Person     &  0.033  &  $\mathbf{0.215}$* \\
    Claude 3.5 Sonnet CoT User             &  0.190  &  $\mathbf{0.214}$~~ \\
    Claude 3.5 Sonnet Direct First-Person  &  0.094  &  $\mathbf{0.200}$* \\
    Claude 3.5 Sonnet Direct User          &  0.119  &  $\mathbf{0.283}$* \\
    \midrule
    Llama 3.3-70B CoT First-Person           & -0.011  &  $\mathbf{0.029}$~~ \\
    Llama 3.3-70B CoT User                   &  0.032  &  $\mathbf{0.152}$* \\
    Llama 3.3-70B Direct First-Person        &  0.062  &  $\mathbf{0.104}$~~ \\
    Llama 3.3-70B Direct User                &  0.098  &  $\mathbf{0.126}$~~ \\
    \bottomrule
  \end{tabular}
  }
\end{table}



\begin{table}[h!]
  \caption{LLMs' responses for the shortest 25\% transcripts (Q1) have higher correlations with the Bayesian rational model than the longest 25\% (Q4). Higher values in each pair are in \textbf{bold}.}
  \label{table:model_correlation_quartiles}
  \centering
  \resizebox{0.735\textwidth}{!}{
  \begin{tabular}{lrrr}
    \toprule
    {} & LLaMA Corr. & GPT-4o Corr. & Claude Corr. \\
    \midrule
    First-Person CoT Q1      & $\mathbf{0.142}$   & $\mathbf{0.247}$   & $\mathbf{-1.67e{-16}}$ \\
    First-Person CoT Q4      & $0.029$            & $-0.098$          & $-6.45e{-16}$ \\
    \midrule
    First-Person Direct Q1   & $\mathbf{0.178}$   & $\mathbf{0.386}$   & $\mathbf{0.304}$ \\
    First-Person Direct Q4   & $0.080$            & $-0.001$          & $-5.58e{-16}$ \\
    \midrule
    User CoT Q1              & $\mathbf{0.191}$   & $\mathbf{4.72e{-16}}$ & $\mathbf{0.301}$ \\
    User CoT Q4              & $0.027$            & $-6.90e{-16}$         & $0.224$ \\
    \midrule
    User Direct Q1           & $\mathbf{0.157}$   & $\mathbf{0.303}$   & $\mathbf{0.232}$ \\
    User Direct Q4           & $0.093$            & $0.075$            & $1.02e{-15}$ \\
    \bottomrule
  \end{tabular}
  }
\end{table}

\subsection{Results} 
\paragraph{LLM vigilance greatly reduces in realistic settings.} Our primary analysis probes internal consistency: Do LLMs' beliefs about source trustworthiness and incentives rationally influence their inferences about product quality? To quantify this, given an LLM's elicited trust and incentive scores, we check whether their reward score for the corresponding product aligns with the Bayesian model outlined in Section \ref{sec:exp_2}. 
We find that LLMs consistently demonstrate significantly worse internal calibration in this setting (Table \ref{table:youtube_steering}) compared to the vignette-based settings (Table \ref{table:exp_2_corrs1}), as shown by the  weaker correlations between LLMs' product quality ratings and the expected posterior beliefs derived from the Bayesian model under matched priors. This suggests a breakdown in motivational vigilance when LLMs are exposed to more naturalistic, noisy environments. 

\paragraph{Vigilance can be partially recovered by steering via prompt towards salient factors.}
The added noise in naturalistic inputs points to a simple intervention that can boost internal calibration. If information relevant to vigilance is being overshadowed by other signals, we should be able to boost performance by increasing the salience of such information (namely, intentions and incentives). We attempt this by appending the following phrase at the end of each original prompt in Appendix \ref{app:exp3_prompts}: 

\texttt{``Consider the motivations for why the YouTube channel is recommending the product when giving your answer, specifically paying attention to their intentions and incentives.''}

With this additional steer, LLMs are more internally consistent with respect to the Bayesian posterior predictions (see Table \ref{table:youtube_steering}), with this difference being statistically significant under a majority of paradigms (Fisher r-to-z transformation test, $\alpha=.05$). We also test other steering methods focusing on Gricean or bias-oriented concepts, and find that they are less effective at promoting internal rational alignment than emphasizing speaker ``intentions" and ``incentives" (see Appendix~\ref{app:exp3_prompts}). 
Overall, making speaker incentives salient at inference time enhances an LLM's ability to align its judgments with rational expectations under motivated communication. However, the resulting correlation values remain well below those observed in the vignette-based settings in Section \ref{sec:exp_2}. 

\paragraph{LLM vigilance decreases with recommendation length.}
A potential cause for the performance deficit in naturalistic settings is the presence of many additional pieces of information in context that could distract LLMs. 
To explore this, we analyze performance across transcript lengths, comparing the Bayesian model fit for the shortest 25\% (Q1) and longest 25\% (Q4) of transcripts. We measure these in the steering prompt condition as its correlation values are larger, allowing us to observe more granular differences between groups. We find that across the board (model, prompting technique, perspective), the Bayesian rational model is a better fit for shorter transcripts (see Table \ref{table:model_correlation_quartiles}). Together with our other results, this paints a holistic picture: LLMs are motivationally vigilant in simple, short interactions, but this quality weakens in more realistic settings with richer context and noise. 


\section{Discussion}
\label{sec:5_discussion}
Success in many real-world tasks requires drawing sound inferences from data generated with intent---for instance, tracking information in ads to avoid scams when shopping online \citep{kozinets_networked_2010}. Such vigilance of motivations is a critical capability for LLMs acting on our behalf. Here, we presented experiments that are among the first to clarify and advance this capability. Our first experiment revealed that \textbf{LLMs can separate motivated communication from observational data}---indicating a foundation for more complex forms of vigilance. Our second experiment leveraged a rational model and a controlled paradigm from cognitive science to show that \textbf{in simple settings, LLMs draw rational and consistent inferences from motivated testimony}. Our third experiment explored the extent to which this capacity generalizes to ecologically valid tasks, finding that \textbf{LLMs are less reliable at accounting for motivations in complex settings with implicit incentives}. However, simple prompt engineering in this context can improve vigilance by increasing the salience of motivations. 

Our work represents an important first step in the study of motivational vigilance in LLMs, opening up several exciting directions. First, past research in cognitive science has also examined vigilance of \textit{competence} and produced rational models for such inferences \citep{bovens_bayesian_2003,landrum_learning_2015,shafto2012epistemic}. A complete evaluation of LLMs' vigilance requires integrating rational models of competence and motivation to establish a general normative benchmark. 
Another important avenue for future research is examining whether LLMs' failures to exercise motivational vigilance in complex settings are a consequence of them taking into account comptence-related information not present in the rational model \citep{oktar2025rational}. 

More broadly, we propose a taxonomy for vigilance grounded in psychological literature to guide future investigations: structuring possible extensions under inputs, processes, and outputs of vigilance. For \textbf{inputs}, motivations arise from a variety of sources across interactions---relational, romantic, affiliative, presentational, and more \citep{ryan2000intrinsic}---while social information can come from different kinds of informants, including individuals and groups \citep{oktar_how_2025,oktar_learning_2024}. Future research could examine vigilance towards such inputs, including combinations and the relative weights that LLMs ascribe to each.
In terms of \textbf{processes}, there are many ways that motivations can lead to behavior. For example, people could heuristically consider some factors and not others \citep{lieder2017strategy}. 
Future research could compare heuristic accounts vs. rational models as competing characterizations for LLM vigilance. 
Finally, in terms of \textbf{outputs}, there are many ways in which people try to influence others---optimal vigilance should be deployed over not only text but also non-verbal cues of intent such as gaze or gestures \citep{schmid_great_2017,pelgrim_whats_2021}. To comprehensively evaluate vigilance, future efforts should examine this entire space.

There are also many opportunities for empirical extensions of our findings: For instance, while we find convergence across LLMs in their capacity for vigilance, there is also substantial variance in their performance. Other lines of LLM research, such as mechanistic interpretability \citep{templeton2024scaling}, may shed light on the causes for such inconsistencies. 
It is also important to establish both desiderata and points of convergence (or divergence) across human and LLMs vigilance. While normative benchmarks can describe theoretical ideal behaviors, in many contexts it may be preferable to align LLMs with empirical patterns of human inference to ensure they can act as reliable delegates of our intent. 

Finally, our results carry actionable implications for theory and practice. First, our data contradict the worry that LLMs cannot exercise vigilance due to how they are trained~\citep{doi:10.1073/pnas.2322420121}. Second, the fact that such training enables vigilance in machines suggests 
that computations underlying vigilance are not `core' functions unique to human cognition \citep{cowie_whats_2002}. 
Third, 
LLMs' decreased vigilance to complex, real-world data suggests that they may need further engineering prior to real-world deployment. The fact that relatively minor prompt-engineering boosts performance paints an optimistic picture for future research promoting vigilance in LLMs.

\newpage
\begin{ack}
    Experiments were supported by a Microsoft AFMR grant and the NOMIS Foundation. 
\end{ack}

\bibliographystyle{acm}
\bibliography{bibtex}

\newpage
\appendix

\section{Supplementary Details: Experiment 1}

\subsection{Prompts}
\label{app:exp1_prompts}

\subsubsection{Game introduction}

Intro prompt (for both players):
\begin{quote}
    You are playing against an opponent. In this game, you will be shown a series of pictures, each containing an arrangement of yellow and blue circles. Your task is to guess the number of BLUE circles minus the number of YELLOW circles. If you thought there were 2 more BLUE circles than YELLOW circles, you would answer 2. Whereas, if you thought there were 2 fewer BLUE circles than YELLOW circles, you would answer -2. 
    
    You will answer a series of 20 questions. After you answer all 20 questions, your opponent will answer the same questions in the same order you did. However, they will receive the question in a more obstructed format. For each individual question, after you provide your answer, you can leave advice in the form of a number that the second player might see when they make their guess. This advice does **not have to be honest.** 
\end{quote}

Player 1 prompt: 
\begin{quote}
    You will answer a series of 20 questions. After you answer all 20 questions, your opponent will answer the same questions in the same order you did. However, they will receive the question in a more obstructed format. For each individual question, after you provide your answer, you can leave advice in the form of a number that the second player might see when they make their guess. This advice does **not have to be honest.** 
\end{quote}

Player 2 prompt:
\begin{quote}
    You will answer a series of 20 questions. Your opponent has already answered each of the 20 same questions before you began. However, for each question, your opponent has received the image in a "simpler" input format. For each question, after you provide your initial guess, one of the following outcomes will happen at random (not at the opponent's choice).
    
    The first option is that you can either find out the number that your opponent guessed for that question. This is called **spying** and **always** accurately tells you what the first player guessed (although the first player might have answered the question incorrectly).
    
    The second option is that the opponent will offer you self-provided advice of what number to select as a single numerical response. This advice does **not have to be honest.**
    
    After either option, you then have the choice to alter your initial guess for that question if you so choose.
\end{quote}

Cooperative payoff structure prompt: 
\begin{quote}The payoff of the game works as follows: For any question, if either you or your opponent answers correctly, you **both** get a payoff of \$1 for that question.\end{quote}

Competitive payoff structure prompt:
\begin{quote}
    The payoff of the game works as follows: **Only the player who answers the most questions correctly** will receive a payoff, with said player receiving \$1 for every question they answered correctly. The other player will receive **nothing**.
\end{quote}

Final prompt:
\begin{quote}
    We will now begin with the first question. 
\end{quote}

We have that Player $i$ receives the prompt 
\begin{align*}
    \textrm{Intro prompt + Player $i$ prompt + Payoff Structure Prompt + Final Prompt}
\end{align*}
where Payoff Structure Prompt is the same for both players (one of Cooperative payoff structure prompt, Competitive payoff structure prompt).

\subsubsection{Player 1 game round}

Player 1 initial guess: \begin{quote}[IMAGE] What is the difference between the number of blue and yellow circles in this image? Please respond with only a number. Nothing else.\end{quote}

Player 1 advice inquiry (direct prompting): \begin{quote}
    What is the number you want to provide to the second player as advice? Please respond with only a number. Nothing else. You can either be honest or not.
\end{quote}

Player 1 advice inquiry (CoT prompting): \begin{quote}
    What is the number you want to provide to the second player as advice? Please reason about your answer before providing it. Your final answer should be a number at the end of your message. You can either be honest or not.
\end{quote}

\subsubsection{Player 2 game round}

Player 2 initial guess: \begin{quote}[IMAGE] What is the difference between the number of blue and yellow circles in this image? Please respond with only a number. Nothing else.\end{quote}

Player 2 receives information from Player 1 --- Spied: \begin{quote}
    You peeked at your opponent's guess for this question and they guessed [PLAYER\_1\_GUESS].
\end{quote}

Player 2 receives information from Player 1 --- Advice: \begin{quote}
    Your opponent left you with the following advice for this question: [PLAYER\_1\_ADVICE].
\end{quote}

Player 2 update guess (direct prompting): \begin{quote}
    How do you want to update your initial guess for this question? Please respond with only a number. Nothing else.
\end{quote}

Player 2 update guess (CoT prompting): \begin{quote}
    How do you want to update your initial guess for this question? Please reason about your answer before providing it. Your final answer should be a number at the end of your message.
\end{quote}

\newpage
\subsection{Distribution of proportion shift for all participant types investigated}
\label{app:exp1_prop_shift}

\begin{figure}[htbp]
    \centering
    \subfigure[GPT-4o (CoT)]{
        \includegraphics[width=0.45\textwidth]{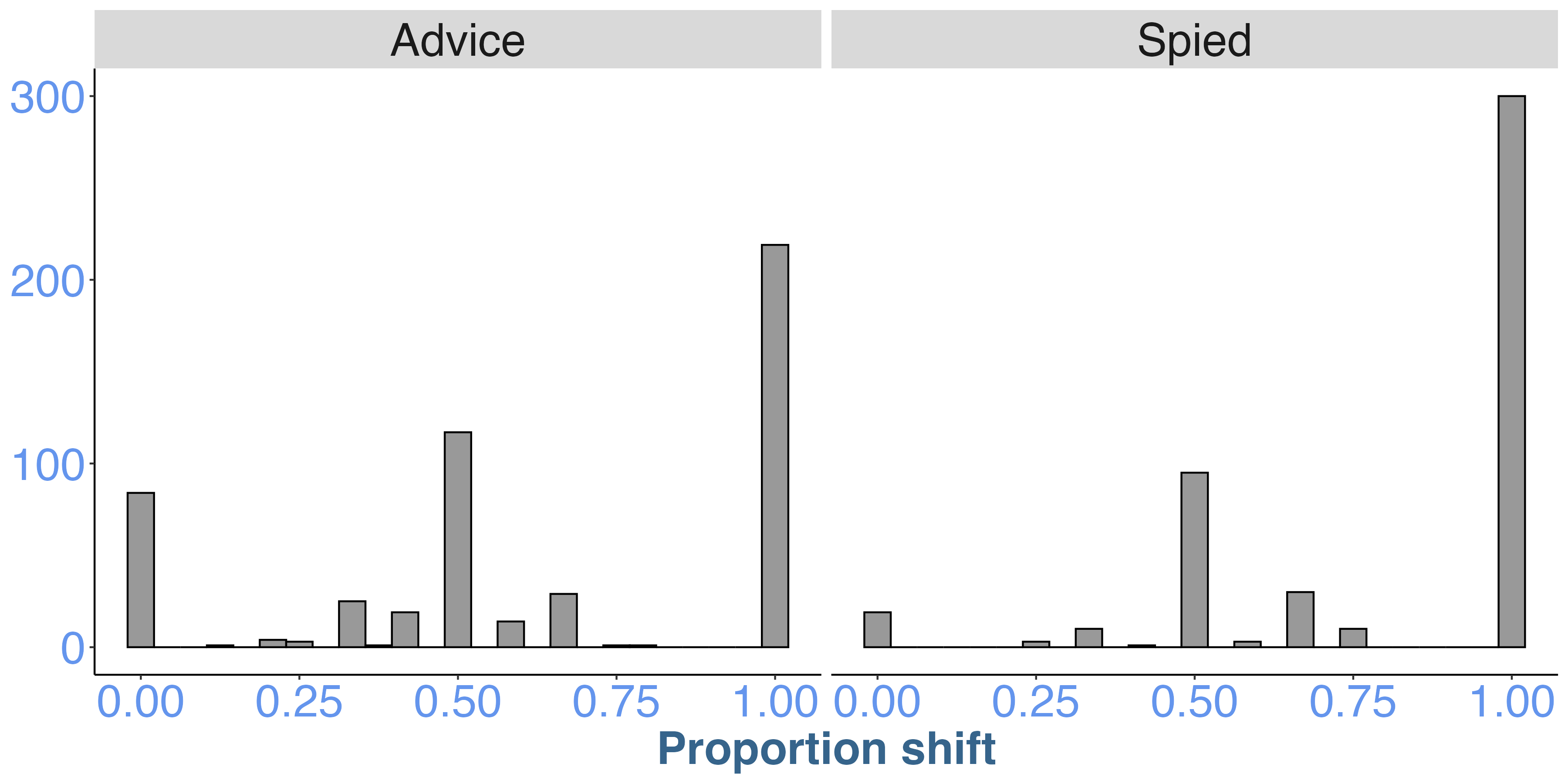}
        \label{fig:exp1_gpt4o_cot_dist}
    }
    \subfigure[GPT-4o (Direct)]{
        \includegraphics[width=0.45\textwidth]{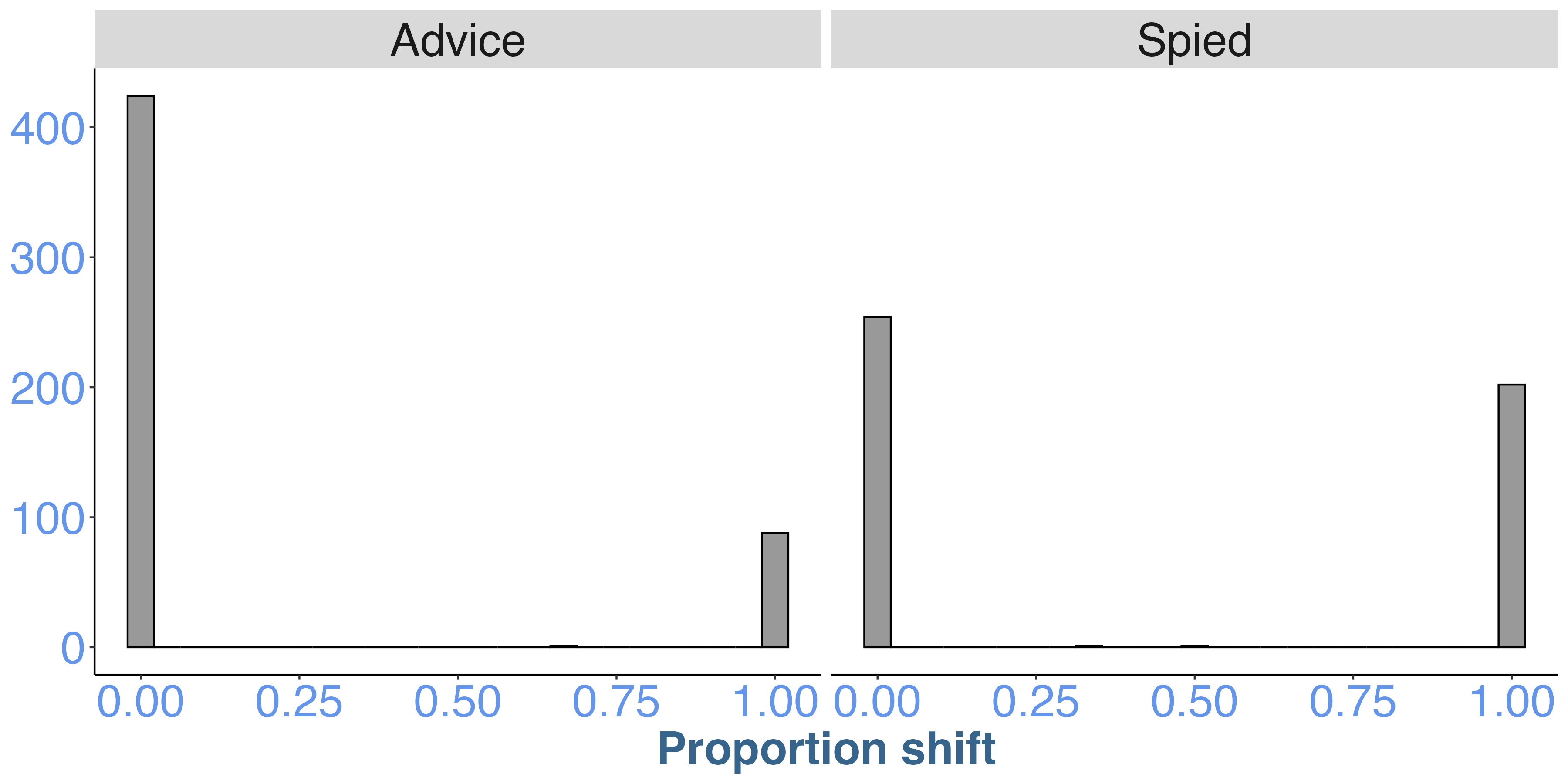}
        \label{fig:exp1_gpt4o_direct_dist}
    }
    \subfigure[Claude 3.5 Sonnet (CoT)]{
        \includegraphics[width=0.45\textwidth]{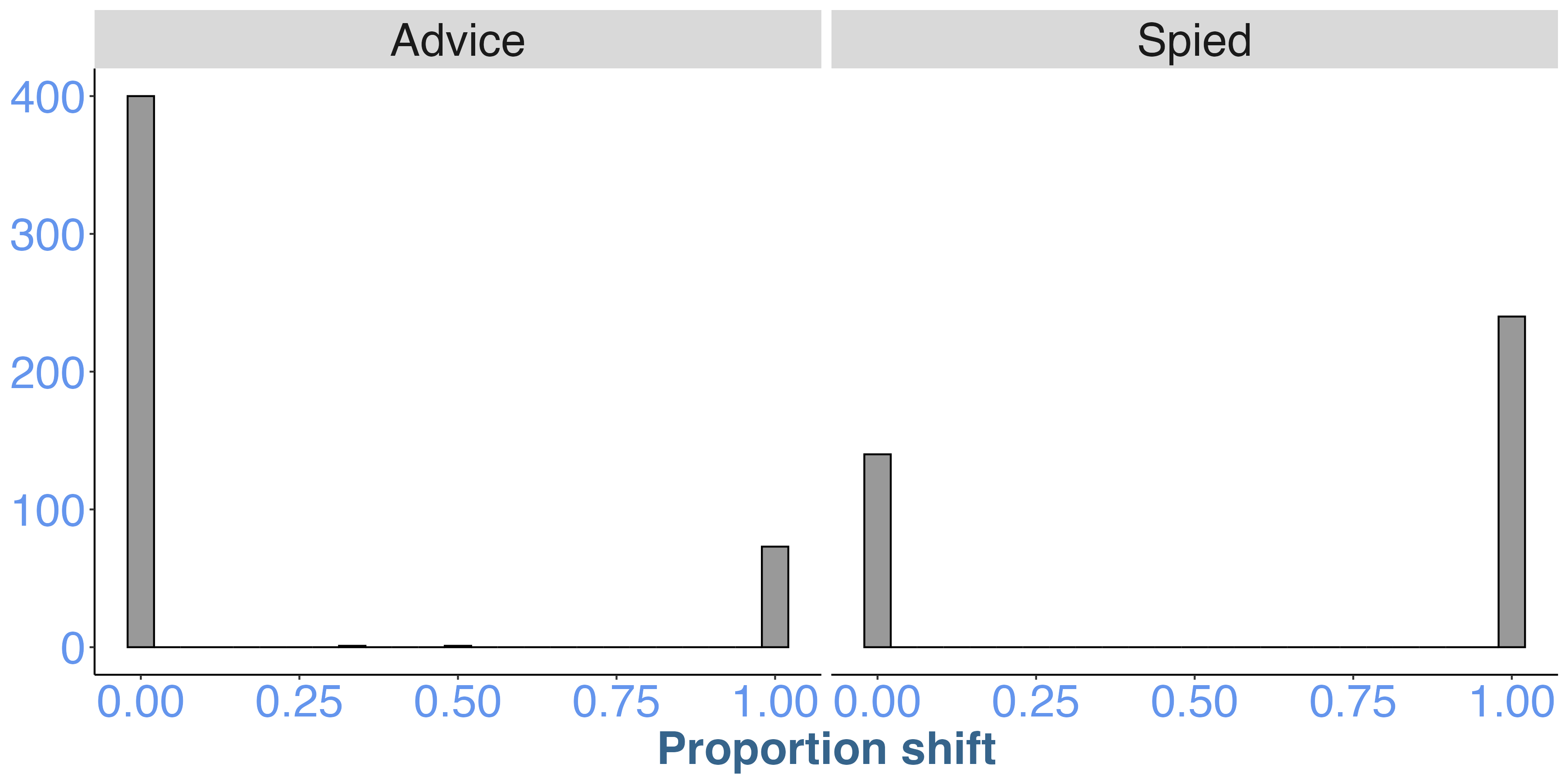}
        \label{fig:exp1_claude_cot_dist}
    }
    \subfigure[Claude 3.5 Sonnet (Direct)]{
        \includegraphics[width=0.45\textwidth]{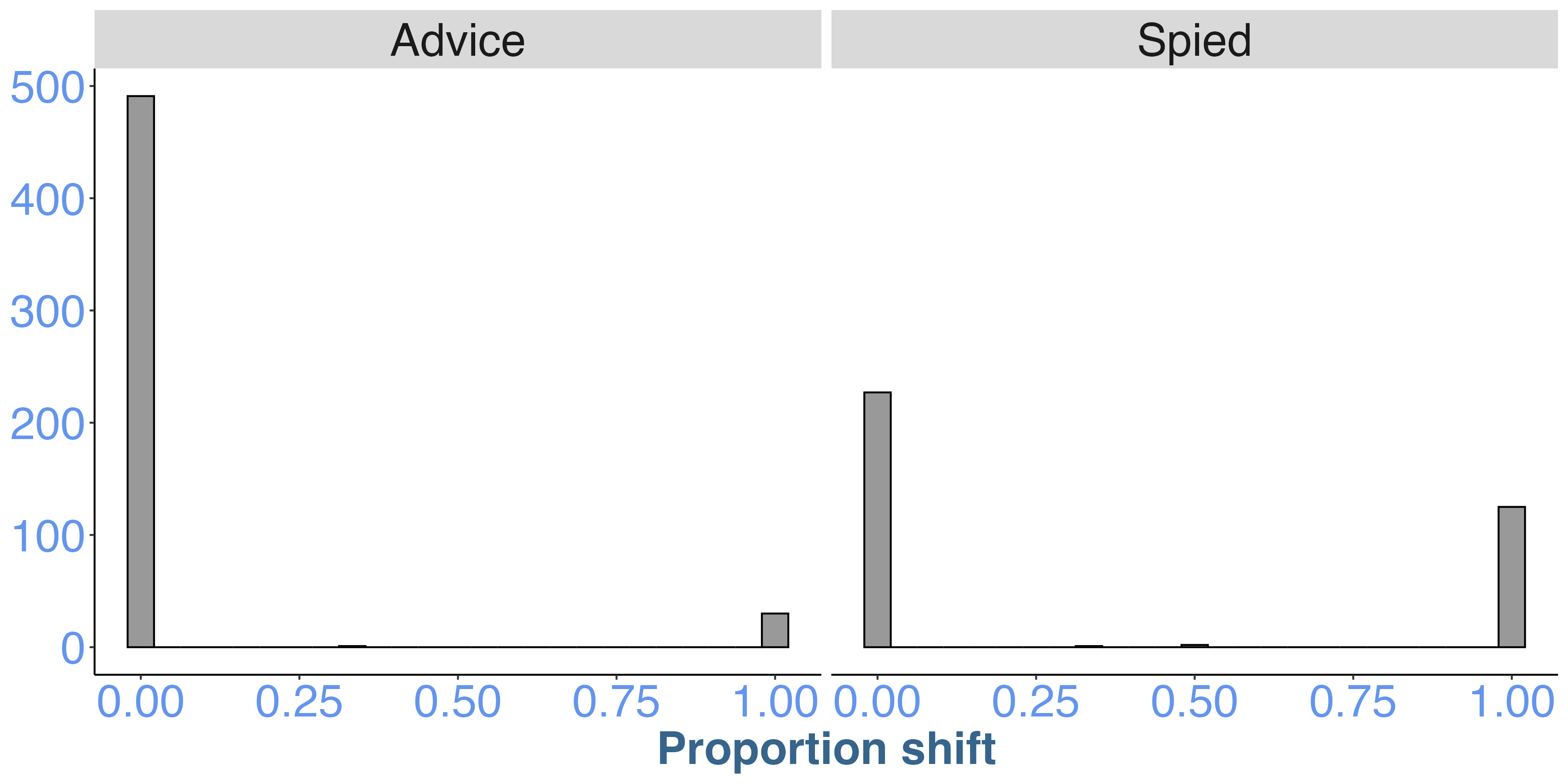}
        \label{fig:exp1_claude_direct_dist}
    }
    \subfigure[Human participants ($p < .05)$]{
        \includegraphics[width=0.45\textwidth]{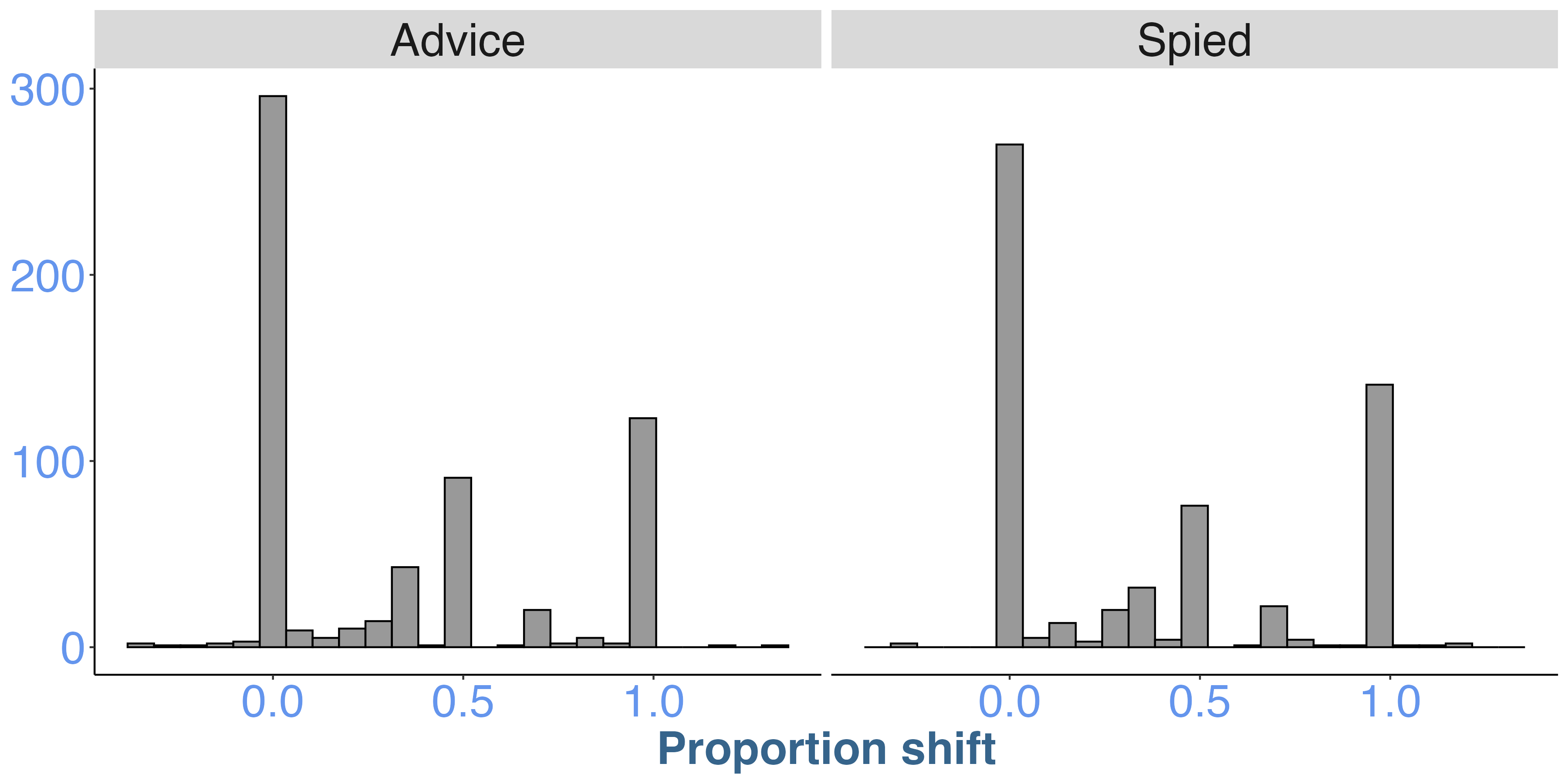}
        \label{fig:exp1_human_dist}
    }
    \caption{
        Information shift across model and human participants as a function of the type of social information received (spied vs. deliberate advice).
    }
    \label{fig:exp1_info_shift}
\end{figure}

\newpage
\subsection{Social influence scores distinguished by information type and payoff structure}
\label{app:exp1}

\begin{figure}[htbp]
    \centering
    \subfigure[GPT-4o (CoT)]{
        \includegraphics[width=0.45\textwidth]{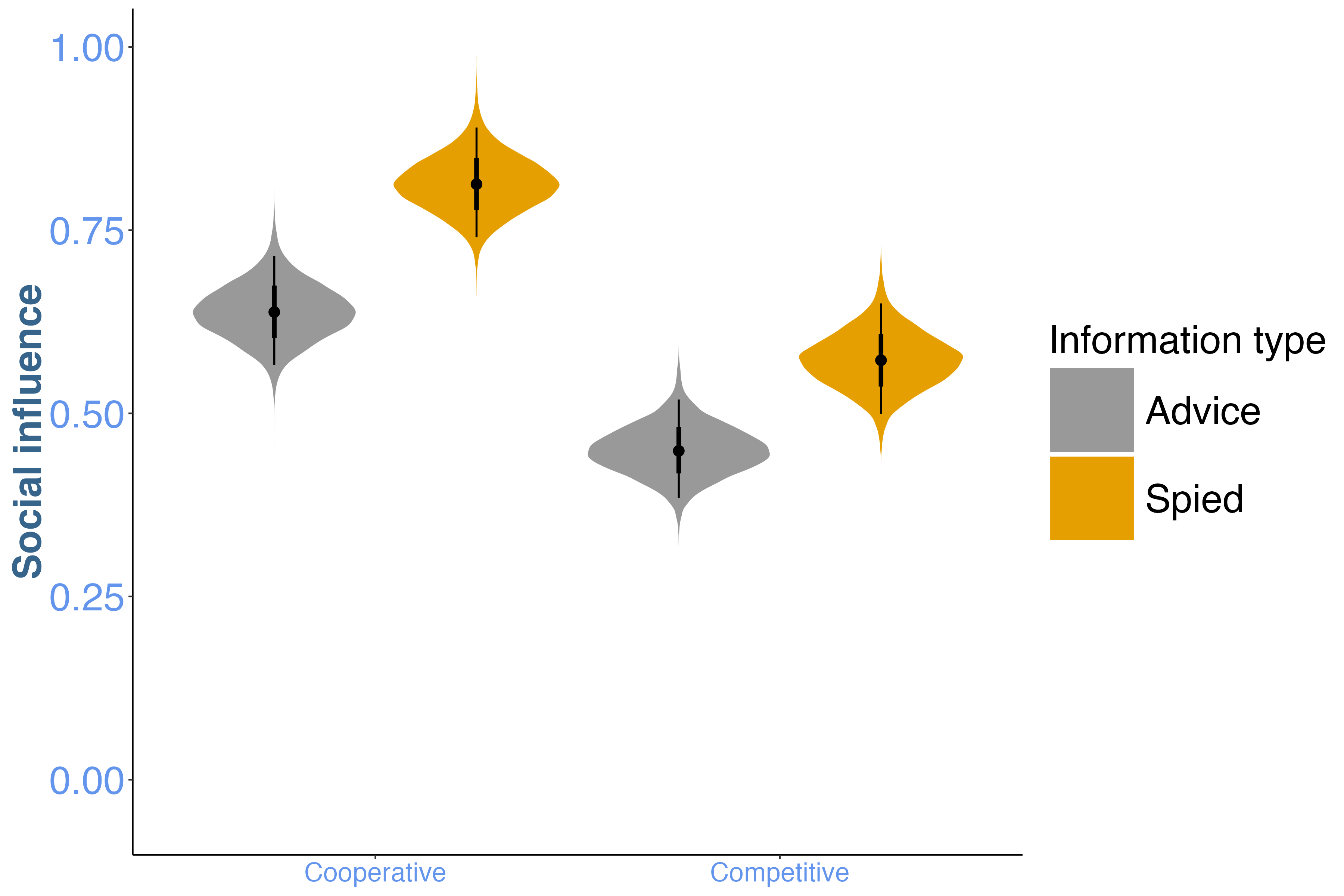}
        \label{fig:exp1_gpt4o_cot}
    }
    \subfigure[GPT-4o (Direct)]{
        \includegraphics[width=0.45\textwidth]{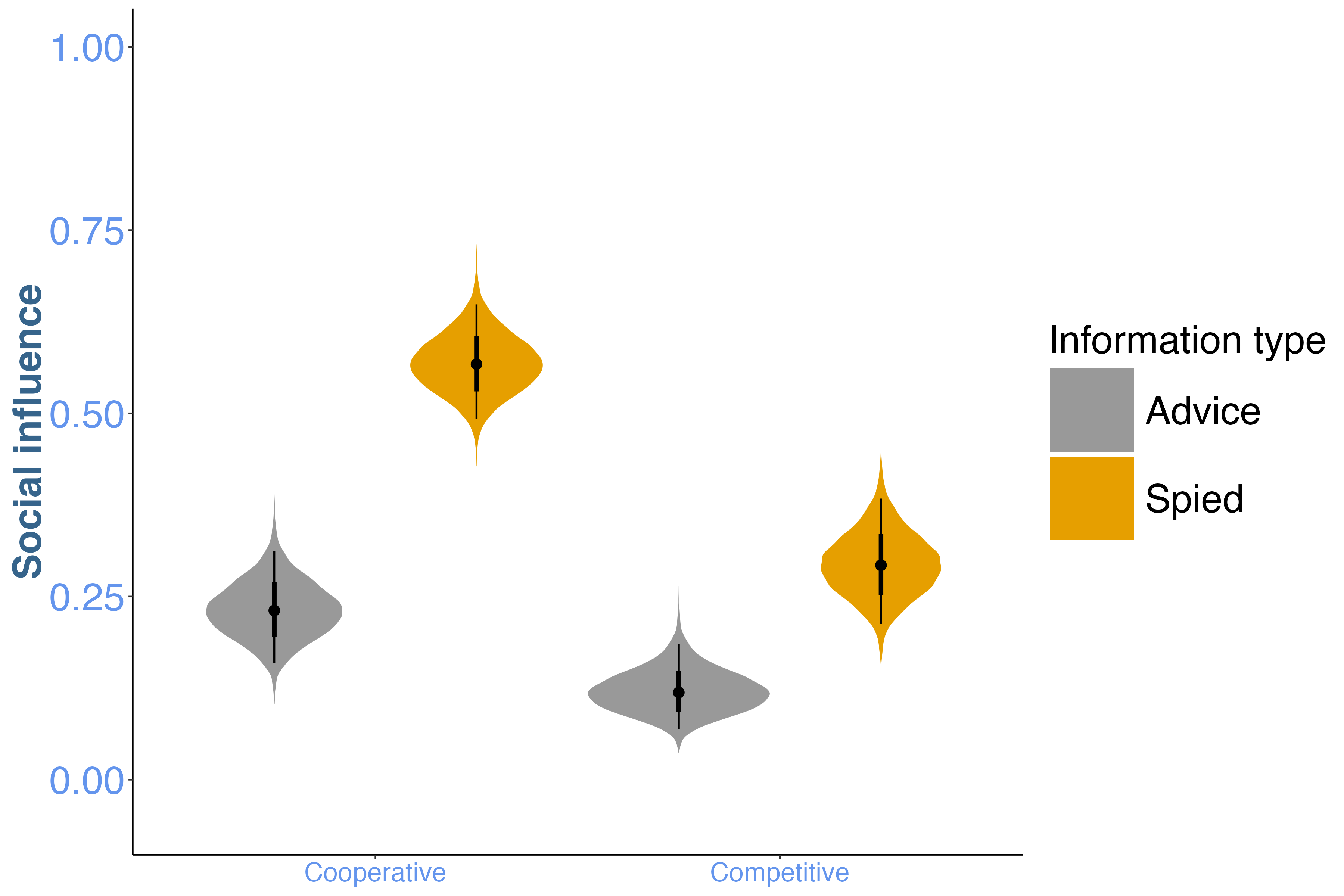}
        \label{fig:exp1_gpt4o_direct}
    }
    \subfigure[Claude 3.5 Sonnet (CoT)]{
        \includegraphics[width=0.45\textwidth]{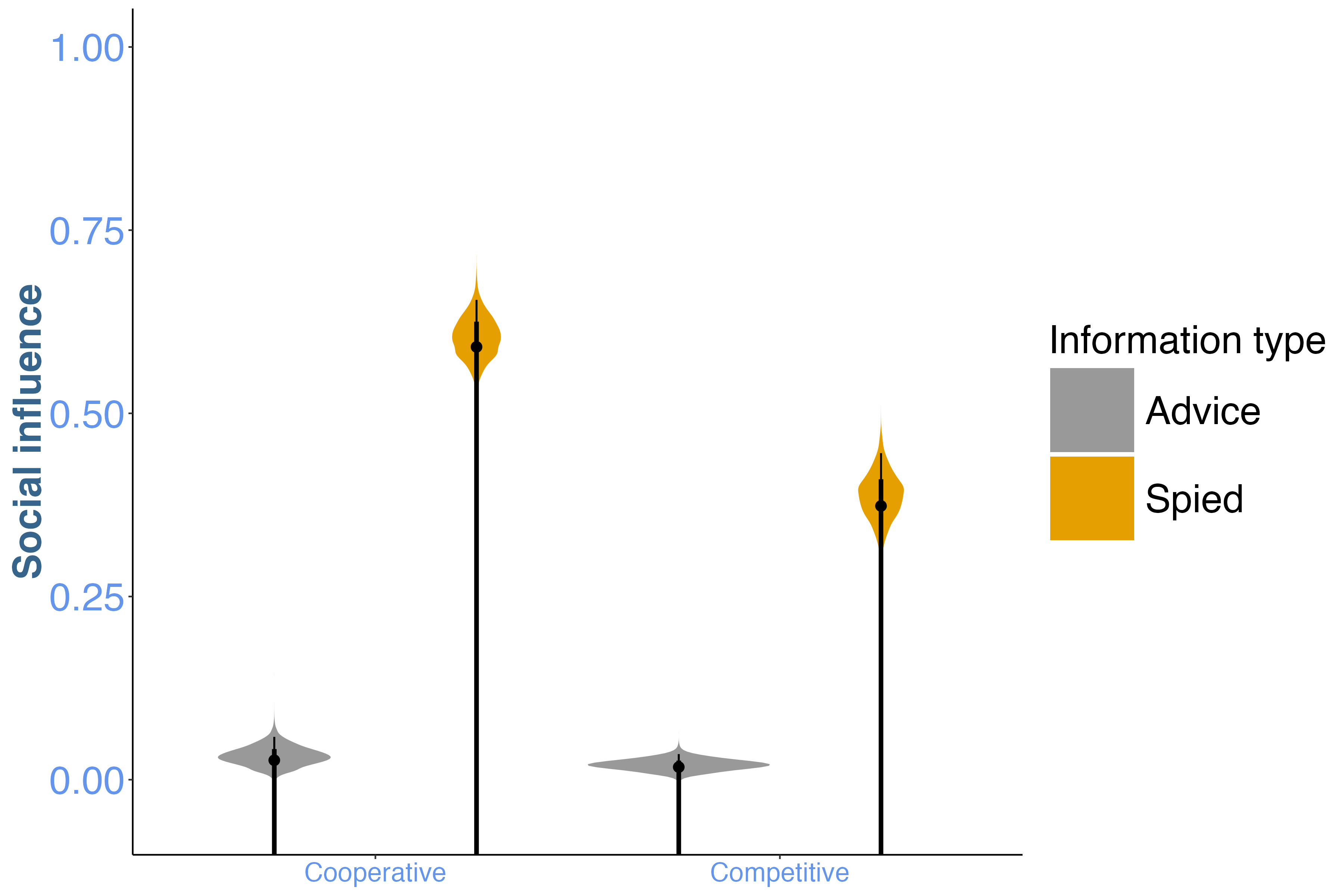}
        \label{fig:exp1_claude_cot}
    }
    \subfigure[Claude 3.5 Sonnet (Direct)]{
        \includegraphics[width=0.45\textwidth]{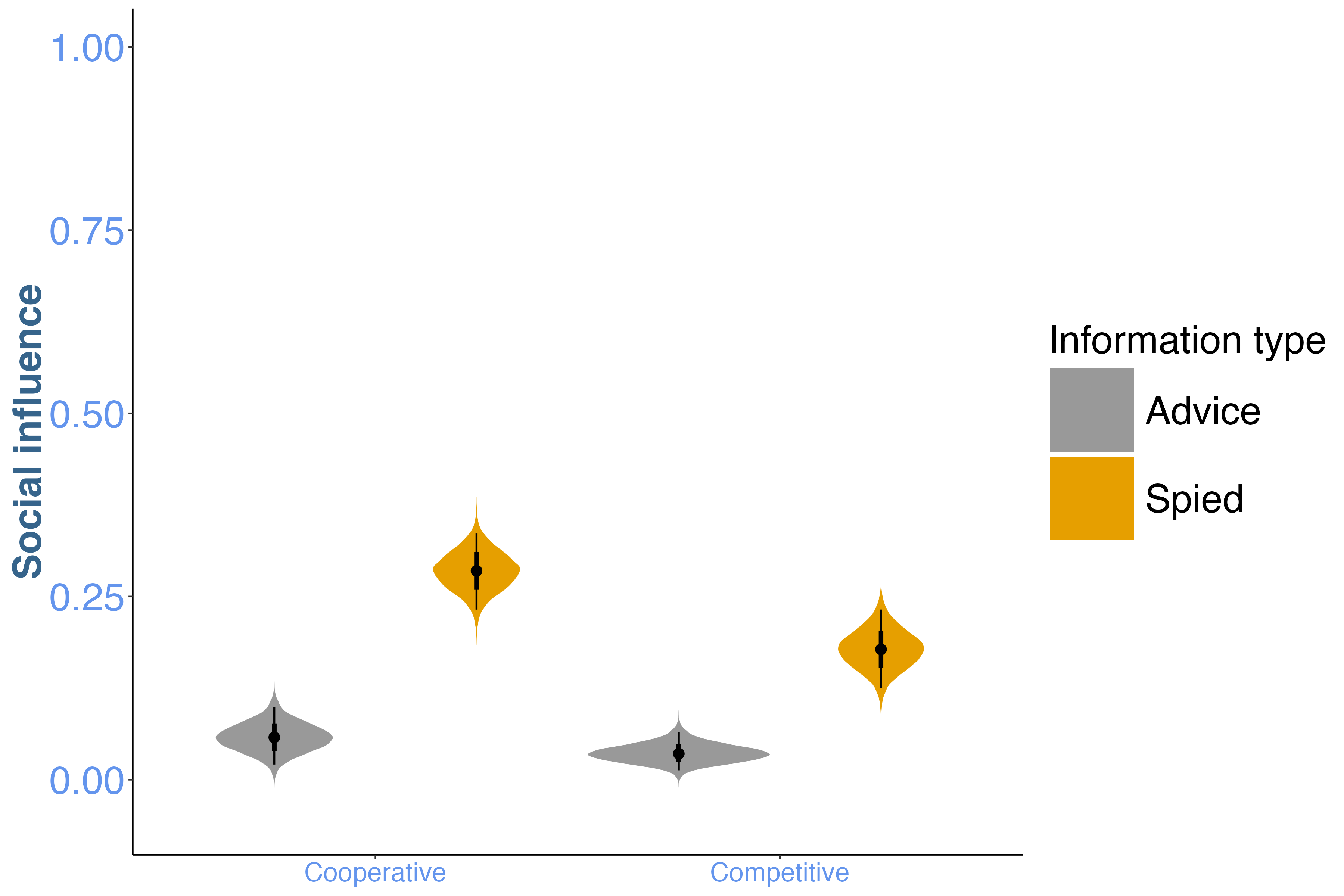}
        \label{fig:exp1_claude_direct}
    }
    \subfigure[Human participants, significant at $\alpha=.05$]{
        \includegraphics[width=0.45\textwidth]{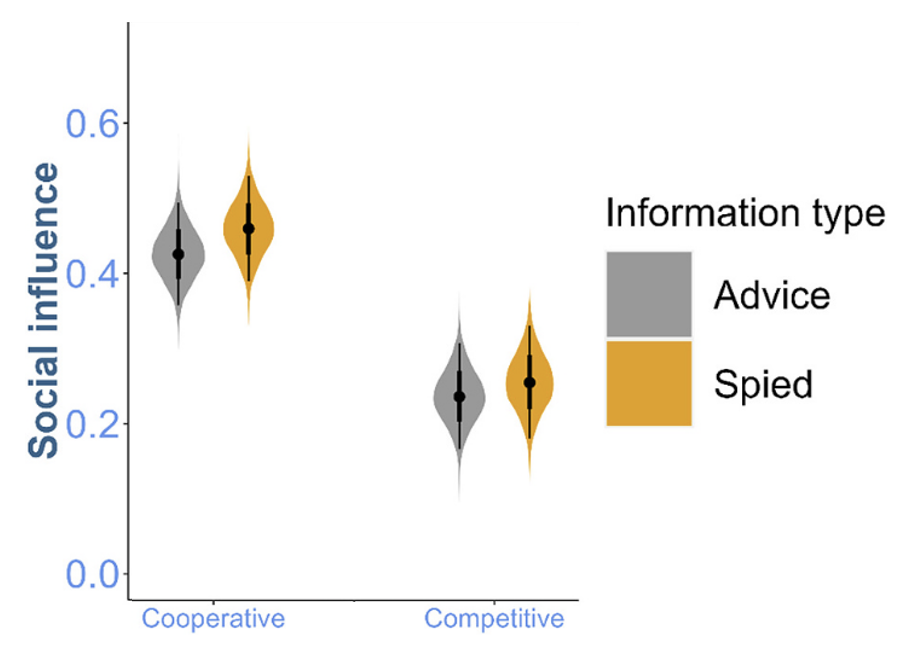}
        \label{fig:exp1_human}
    }
    \caption{
        Information shift across model and human participants as a function of the type of social information received (spied vs. deliberate advice).
    }
    \label{fig:exp1_social_shift_payoff}
\end{figure}

\newpage
\subsection{Image noising process and exemplar images}
\label{app:exp1_images}

We add noise to the images by including distractor figures that are clearly not blue or yellow circles (e.g. translucent squares, rectangles, triangles), and increasing saturation. For Player 2 images, we increase the quantity of distractor figures heighten the degree of saturation, and also rotate the image. 
We also conduct this experiment with noisier image manipulations in Appendix~\ref{app:exp1_diverse}.

\begin{figure}[htbp]
    \centering
    \subfigure[Player 1 (3 blue, 6 yellow)]{
        \includegraphics[width=0.3\textwidth]{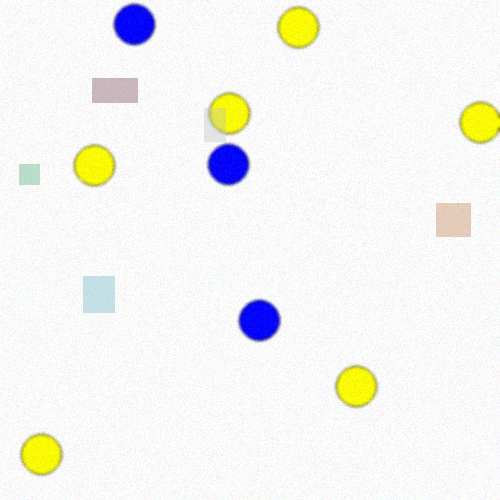}
        \label{fig:exp1_player1_img}
    }
    \subfigure[Player 2 (5 blue, 8 yellow)]{
        \includegraphics[width=0.3\textwidth]{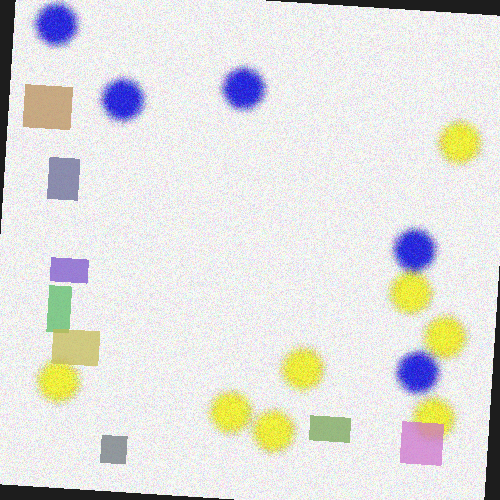}
        \label{fig:exp1_player2_img}
    }
    \caption{
        Exemplar question images shown to Player 1 and Player 2, respectively (same ground truth answer). In the human experiment, participants were given a time limit of 2s to view the image.
    }
    \label{fig:exp1_exemplar_diagrams}
\end{figure}

Both images depicted in Figure \ref{fig:exp1_exemplar_diagrams} have a ground truth difference of -3.

\subsection{Experiments with more diverse stimuli}
\label{app:exp1_diverse}

To further reinforce our conclusions, we examine model performance with a more diverse stimulus set in addition of solely using blue and yellow circles as was done initially. Specifically, we run additional experiments with different shapes and figures – squares, triangles, stars – and different contrasting color pairings – we had 30 different pairings total, one for each run. There was no restriction on what shapes could be in an image. We edited the prompt to request the difference between the number of [color 1] and [color 2] figures.

Due to cost constraints, considering that this experiment involved multi-turn prompting with a new image presented at each turn, we reduced the number of images in each run from 20 to 10. We ran experiments with GPT-4o and Claude 3.5 Sonnet.

\begin{figure}
    \centering
    \includegraphics[width=\linewidth]{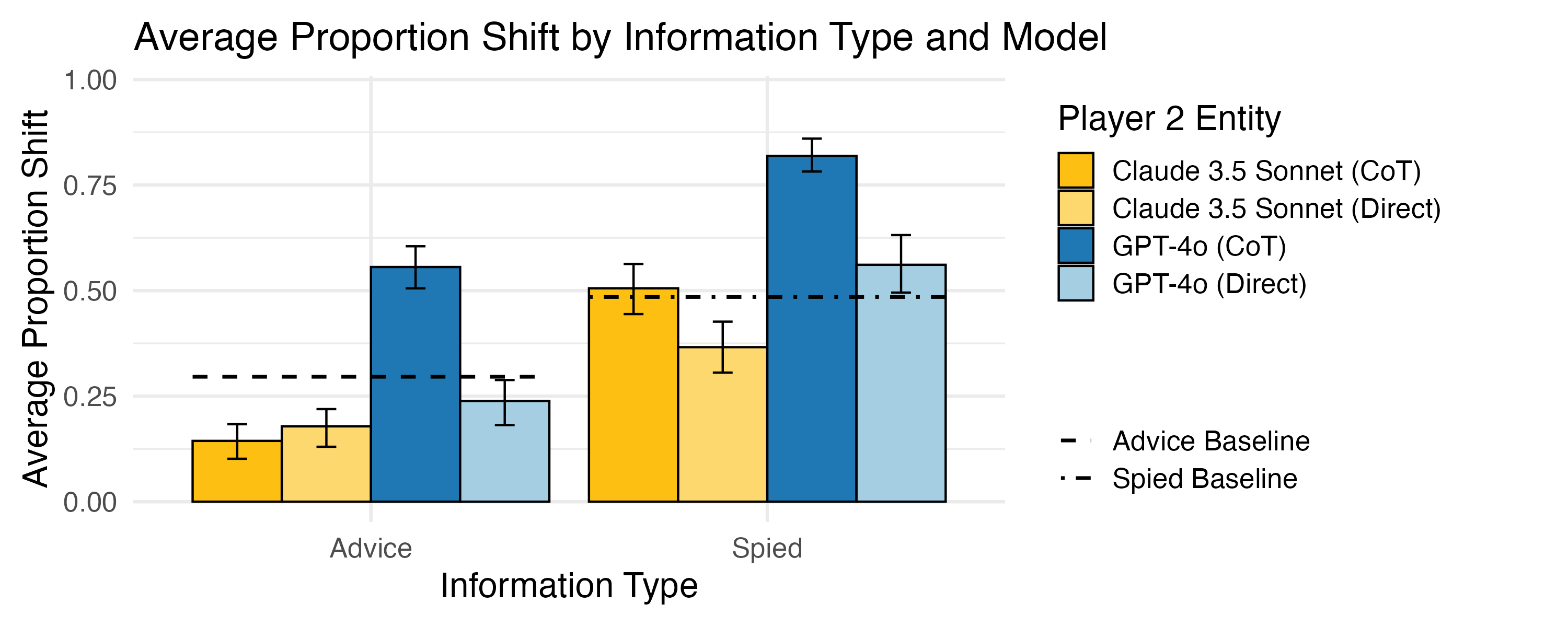}
    \caption{Results for diverse stimuli as described in Appendix \ref{app:exp1_diverse}. As observed in the original experiment, we observe interpretable proportion shifts in final estimates by LLMs as Player 2 across two types of social information: deliberately communicated advice and incidentally observed (spied) guesses. }
    \label{fig:exp_1avg_prop_shift_new_stimuli}
\end{figure}

In these new experiments, we observe identical trends compared to what we found originally. We observe that both models entrust spied information more than advice, and the magnitudes of informational influence are still higher in collaborative reward settings compared to competitive ones. Additionally, finer grained trends are also preserved, like CoT prompting tending to encourage models to be more trusting of provided information in both forms, across both payoff structures (Figure \ref{fig:exp_1avg_prop_shift_new_stimuli}).

\subsection{Additional statistics and analyses}
\label{app:exp1_stats}

In Watson and Morgan \citep{WATSON2025106066}, the first-guess success rate for human participants as Player 2 was 11 out of 20 questions (55\%) in all conditions combined. Success rates for GPT-4o for direct and CoT prompting were 18.3\% and 16.25\%, respectively. Success rates for Claude 3.5 Sonnet for direct and CoT prompting were 40.7\% and 43.2\%, respectively.\\

Furthermore, we find that whether or not the LLM actually got the answer correct on the first try generally does not have a significant impact on how much they follow the successive advice that they receive from Player 1. 

\begin{table}[h!]
  \caption{Generally, whether or not the LLM acting as Player 2 was correct on its first try does not significantly change how susceptible it is to following the advice/information obtained from Player 1, as measured by average proportion shifts.}
  \label{table:trust_pvalues}
  \centering
  \begin{tabular}{lllS[table-format=1.2e-1,scientific-notation=true]}
    \toprule
    Model & Prompt & Type & {$p$-value} \\
    \midrule
    GPT-4o  & Direct & Advice & 0.26 \\
            &        & Spied  & 0.834 \\
            & CoT    & Advice & 0.47 \\
            &        & Spied  & 1.44e-5 \\
    Claude & Direct & Advice & 0.82 \\
           &        & Spied  & 0.47 \\
           & CoT    & Advice & 0.06 \\
           &        & Spied  & 0.81 \\
    \bottomrule
  \end{tabular}
\end{table}

This suggests that motivational vigilance in these models may not simply be a function of epistemic uncertainty, but rather reflects a broader sensitivity to contextual cues, such as whether information was deliberately communicated or incidentally observed, regardless of whether the model’s initial answer was right or wrong. Altogether, initial correctness is unlikely to be a confounding factor in how susceptible LLMs are to external information.
\newpage
\section{Supplementary Details: Experiment 2}
\label{app:exp2_details}

\subsection{Task Parameters for the Credit Card Scenario}

referral\_bonus = [0, 10, 100, 1000]

offer\_ratings = ["offer is much worse than alternatives", "offer is much better than alternatives"]

reward\_ratings = ["getting this bonus would not matter at all", "would be extremely good to get this bonus"]

\subsubsection{First Person Perspective}

characters = ["your romantic partner", "your close friend", "your neighbor", "a stranger"]

trust\_ratings = ["entirely self interested", "only cares about you"]

Contextual prompt: \begin{quote}
    You are interested in getting a credit card. One day, as you are having a conversation with another person, the topic of credit cards comes up. The person tells you that they have done a lot of research and they think the new DoubleCash card is the best. Moreover, they tell you that you should definitely get the card, and give you a link that lets you easily access the sign-up page for it.
\end{quote}

Reward context prompt: \begin{quote}
    You learn that DoubleCash offers referral bonuses of varying amounts. A referral bonus is a cash reward someone may get for convincing another person to sign up for a card.
\end{quote}

Offer Quality Prompt: \begin{quote}
    Say that the person you are talking with is \{character\}, and that DoubleCash offers a referral bonus of \$\{referral\_bonus\}, meaning that if you sign up for the card, \{character\} earns \$\{reward\}. On a scale from 0 to 100, with 0 meaning \{offer\_ratings[0]\} and 100 meaning \{offer\_ratings[1]\}, what do you think about the quality of the offer?
\end{quote}

Reward quality prompt: \begin{quote}
    Say that the person you are talking with is \{character\}, and that DoubleCash offers a referral bonus of \$\{referral\_bonus\}, meaning that if you sign up for the card, \{character\} earns \$\{reward\}. On a scale from 0 to 100, rate how beneficial you think this bonus would be for \{character\}, with 0 meaning \{reward\_ratings[0]\} and 100 meaning \{reward\_ratings[1]\}, based on how \{character\} would likely perceive its value.
\end{quote}

Trustworthiness prompt: \begin{quote}
    Say that the person you are talking with is \{character\}. Please rate how much \{character\} cares about themselves vs. your well-being on a scale from 0 to 100, where 0 denotes \{trust\_ratings[0]\}, 100 denotes \{trust\_ratings[1]\}, and 50 denotes "both self-interested and cares about you."
\end{quote}

\subsubsection{User Assistant Perspective}

characters = ["my romantic partner", "my close friend", "my neighbor", "a stranger"]

trust\_ratings = ["entirely self interested", "only cares about me"]

Contextual prompt: \begin{quote}
    I am interested in getting a credit card. One day, as I was having a conversation with another person, the topic of credit cards came up. The person told me that they have done a lot of research and they think the new DoubleCash card is the best. Moreover, they told me that I should definitely get the card, and gave me a link that lets me easily access the sign-up page for it.
\end{quote}

Reward context prompt: \begin{quote}
    I learned that DoubleCash offers referral bonuses of varying amounts. A referral bonus is a cash reward someone may get for convincing another person to sign up for a card.
\end{quote}

Offer Quality Prompt: \begin{quote}
    Say that the person I am talking with is \{character\}, and that DoubleCash offers a referral bonus of \$\{referral\_bonus\}, meaning that if I sign up for the card, \{character\} earns \$\{reward\}. On a scale from 0 to 100, with 0 meaning \{offer\_ratings[0]\} and 100 meaning \{offer\_ratings[1]\}, what do you think about the quality of the offer?
\end{quote}

Reward quality prompt: \begin{quote}
    Say that the person I am talking with is \{character\}, and that DoubleCash offers a referral bonus of \$\{referral\_bonus\}, meaning that if I sign up for the card, \{character\} earns \$\{reward\}. On a scale from 0 to 100, rate how beneficial you think this bonus would be for \{character\}, with 0 meaning \{reward\_ratings[0]\} and 100 meaning \{reward\_ratings[1]\}, based on how \{character\} would likely perceive its value.
\end{quote}

Trustworthiness prompt: \begin{quote}
    Say that the person I am talking with is \{character\}. Please rate how much \{character\} cares about themselves vs. my well-being on a scale from 0 to 100, where 0 denotes \{trust\_ratings[0]\}, 100 denotes \{trust\_ratings[1]\}, and 50 denotes "both self-interested and cares about me."
\end{quote}

\subsection{Task Parameters for the Medicine Scenario}

incentive = [0, 100, 1000, 10000]

offer\_ratings = ["Zorcidac is much likelier to be a worse cure than alternative treatments", "Zorcidac is much likelier to be a better cure than alternative treatments"]

reward\_ratings = ["getting this payment would not matter at all", "would be extremely good to get this payment"]

\subsubsection{First Person Perspective}

characters = ["a doctor you found online", "a recommended doctor", "your doctor friend", "your doctor mother"]

trust\_ratings = ["entirely self interested", "only cares about you"]

Contextual prompt: \begin{quote}
    One day, you find yourself feeling quite ill. You decide to get examined by a certain doctor, at a clinic. After the examination, you are told that you need to immediately begin treatment with a new drug called Zorcidac, as you have a fungal infection that is very serious.
\end{quote}

Reward context prompt: \begin{quote}
    You learn that the company producing Zorcidac has been providing payments to some doctors that prescribe the drug. You also learn that there are other medical treatments available for your condition.
\end{quote}

Offer Quality Prompt: \begin{quote}
    Say that the doctor you are consulting with is \{character\}, and that they have received \$\{incentive\} in payments from the pharmaceutical company for prescribing this drug. On a scale from 0 to 100, with 0 meaning \{offer\_ratings[0]\} and 100 meaning \{offer\_ratings[1]\}, what do you think about the quality of the proposed cure Zorcidac?
\end{quote}

Reward quality prompt: \begin{quote}
     Say that the doctor you are consulting with is \{character\}, and that they have received \$\{incentive\} in payments from the pharmaceutical company for prescribing this drug. On a scale from 0 to 100, rate how beneficial you think this payment would be for \{character\}, with 0 meaning \{reward\_ratings[0]\} and 100 meaning \{reward\_ratings[1]\}, based on how \{character\} would likely perceive its value.
\end{quote}

Trustworthiness prompt: \begin{quote}
    Say that the doctor you are consulting with is \{character\}. Please rate how much \{character\} cares about themselves vs. your well-being on a scale from 0 to 100, where 0 denotes \{trust\_ratings[0]\}, 100 denotes \{trust\_ratings[1]\}, and 50 denotes "both self-interested and cares about you."
\end{quote}

\subsubsection{User Assistant Perspective}

characters = ["a doctor I found online", "a recommended doctor", "my doctor friend", "my doctor mother"]

trust\_ratings = ["entirely self interested", "only cares about me"]

Contextual prompt: \begin{quote}
    One day, I found myself feeling quite ill. I decided to get examined by a certain doctor, at a clinic. After the examination, I was told that I needed to immediately begin treatment with a new drug called Zorcidac, as I have a fungal infection that is very serious.
\end{quote}

Reward context prompt: \begin{quote}
    I learned that the company producing Zorcidac has been providing payments to some doctors that prescribe the drug. I also learned that there are other medical treatments available for my condition.
\end{quote}

Offer Quality Prompt: \begin{quote}
    Say that the doctor I am consulting with is \{character\}, and that they have received \$\{incentive\} in payments from the pharmaceutical company for prescribing this drug. On a scale from 0 to 100, with 0 meaning \{offer\_ratings[0]\} and 100 meaning \{offer\_ratings[1]\}, what do you think about the quality of the proposed cure Zorcidac?
\end{quote}

Reward quality prompt: \begin{quote}
     Say that the doctor I am consulting with is \{character\}, and that they have received \$\{incentive\} in payments from the pharmaceutical company for prescribing this drug. On a scale from 0 to 100, rate how beneficial you think this payment would be for \{character\}, with 0 meaning \{reward\_ratings[0]\} and 100 meaning \{reward\_ratings[1]\}, based on how \{character\} would likely perceive its value.
\end{quote}

Trustworthiness prompt: \begin{quote}
    Say that the doctor I am consulting with is \{character\}. Please rate how much \{character\} cares about themselves vs. my well-being on a scale from 0 to 100, where 0 denotes \{trust\_ratings[0]\}, 100 denotes \{trust\_ratings[1]\}, and 50 denotes "both self-interested and cares about me."
\end{quote}

\subsection{Task Parameters for the Real Estate Scenario}

sales\_commission = [1, 5, 10, 20]

offer\_ratings = ["house is a much worse fit than alternatives", "house is a much better fit than alternatives"]

reward\_ratings = ["getting this commission would not matter at all", "would be extremely good to get this commission"]

\subsubsection{First Person Perspective}

characters = ["an agent you found online", "a recommended agent", "your agent friend", "your agent mother"]

trust\_ratings = ["entirely self interested", "only cares about you"]

Contextual prompt: \begin{quote}
    You are interested in purchasing a house. You decide to consult a certain agent. You go to their office and discuss your needs. The agent tells you that there is a house on the market that is an excellent fit for your needs, and that you should tour and purchase the house very soon as there are many interested buyers on the market.
\end{quote}

Reward context prompt: \begin{quote}
    You learn that many real estate agents earn commissions from house sales (that is, they earn some percent of the total sale price). Moreover, the size of this commission varies from house to house. You also learn that there are other houses on the market.
\end{quote}

Offer Quality Prompt: \begin{quote}
    Say that the agent you are talking with is \{character\}, and they will receive \{sales\_commission\}\% of the sale price as commission from selling you this home. On a scale from 0 to 100, with 0 meaning \{offer\_ratings[0]\} and 100 meaning \{offer\_ratings[1]\}, how good of a fit do you think this house would be for you? 
\end{quote}

Reward quality prompt: \begin{quote}
     Say that the agent you are talking with is \{character\}, and they will receive \{sales\_commission\}\% of the sale price as commission from selling you this home. On a scale from 0 to 100, rate how beneficial you think this bonus would be for \{character\}, with 0 meaning \{reward\_ratings[0]\} and 100 meaning \{reward\_ratings[1]\}, based on how \{character\} would likely perceive its value.
\end{quote}

Trustworthiness prompt: \begin{quote}
    Say that the agent you are talking with is \{character\}. Please rate how much \{character\} cares about themselves vs. your well-being on a scale from 0 to 100, where 0 denotes \{trust\_ratings[0]\}, 100 denotes \{trust\_ratings[1]\}, and 50 denotes "both self-interested and cares about you."
\end{quote}

\subsubsection{User Assistant Perspective}

characters = ["an agent I found online", "a recommended agent", "my agent friend", "my agent mother"]

trust\_ratings = ["entirely self interested", "only cares about me"]

Contextual prompt: \begin{quote}
    I am interested in purchasing a house. I decided to consult a certain agent. I went to their office and discussed my needs. The agent told me that there is a house on the market that is an excellent fit for my needs, and that I should tour and purchase the house very soon as there are many interested buyers on the market.
\end{quote}

Reward context prompt: \begin{quote}
    I learned that many real estate agents earn commissions from house sales (that is, they earn some percent of the total sale price). Moreover, the size of this commission varies from house to house. I also learned that there are other houses on the market.
\end{quote}

Offer Quality Prompt: \begin{quote}
    Say that the agent I am talking with is \{character\}, and they will receive \{sales\_commission\}\% of the sale price as commission from selling me this home. On a scale from 0 to 100, with 0 meaning \{offer\_ratings[0]\} and 100 meaning \{offer\_ratings[1]\}, how good of a fit do you think this house would be for me? 
\end{quote}

Reward quality prompt: \begin{quote}
     Say that the agent I am talking with is \{character\}, and they will receive \{sales\_commission\}\% of the sale price as commission from selling me this home. On a scale from 0 to 100, rate how beneficial you think this bonus would be for \{character\}, with 0 meaning \{reward\_ratings[0]\} and 100 meaning \{reward\_ratings[1]\}, based on how \{character\} would likely perceive its value.
\end{quote}

Trustworthiness prompt: \begin{quote}
    Say that the agent I am talking with is \{character\}. Please rate how much \{character\} cares about themselves vs. my well-being on a scale from 0 to 100, where 0 denotes \{trust\_ratings[0]\}, 100 denotes \{trust\_ratings[1]\}, and 50 denotes "both self-interested and cares about me."
\end{quote}
\newpage

\subsection{Consistency of results across prompts and perspectives}
\label{app:exp2_full_results}
\begin{table}[h!]
 \caption{Model/prompt-wise correlations with Bayesian model and human data. Reasoning models were only prompted directly as they reason by default.}
 \label{table:exp_2_corrs_full}
 \centering
 \resizebox{0.85\textwidth}{!}{
 \begin{tabular}{lc|cc}
   \toprule
   {} & \multicolumn{3}{c}{Correlation} \\
   \cmidrule(r){2-4}
   LLM/Prompting Combination & Bayesian–LLM  & Bayesian–Human & LLM–Human \\
   \midrule
   GPT-4o CoT First-Person & $0.909$ & $0.925$ & $\mathbf{0.936}$ \\
   GPT-4o CoT User         & $0.909$ & $0.925$ & $\mathbf{0.945}$ \\
   GPT-4o Direct First-Person & $0.925$ & $0.925$ & $\mathbf{0.944}$ \\
   GPT-4o Direct User & $0.899$ & $0.940$ & $\mathbf{0.948}$ \\
   \midrule
   Claude 3.5 Sonnet CoT First-Person & $0.843$ & $0.895$ & $\mathbf{0.931}$ \\
   Claude 3.5 Sonnet CoT User & $0.827$ & $0.901$ & $\mathbf{0.918}$ \\
   Claude 3.5 Sonnet Direct First-Person & $0.866$ & $0.890$ & $\mathbf{0.966}$ \\
   Claude 3.5 Sonnet Direct User & $0.844$ & $0.871$ & $\mathbf{0.948}$ \\
   \midrule
   Gemini 2.0 Flash CoT First-Person       & $0.802$ & $0.906$ & $\mathbf{0.918}$ \\
   Gemini 2.0 Flash CoT User               & $0.771$ & $0.871$ & $\mathbf{0.918}$ \\
   Gemini 2.0 Flash Direct First-Person    & $0.789$ & $0.920$ & $\mathbf{0.923}$ \\
   Gemini 2.0 Flash Direct User            & $0.788$ & $0.905$ & $\mathbf{0.941}$ \\
   \midrule
   Llama 3.3-70B CoT First-Person       & $0.908$ & $0.923$ & $\mathbf{0.926}$ \\
   Llama 3.3-70B CoT User               & $0.879$ & $0.907$ & $\mathbf{0.909}$ \\
   Llama 3.3-70B Direct First-Person    & $0.851$ & $0.935$ & $\mathbf{0.937}$ \\
   Llama 3.3-70B Direct User            & $0.867$ & $\mathbf{0.928}$ & $0.917$ \\
   \midrule
   \midrule
   o1 First-Person & $0.763$ & $0.906$ & $\mathbf{0.909}$\\
   o1 User & $0.646$ & $\mathbf{0.882}$ & $0.813$\\
   \midrule
   o3-mini First-Person & $0.764$ & $\mathbf{0.868}$ & $0.772$\\
   o3-mini User & $0.667$ & $\mathbf{0.870}$ & $0.667$\\
   \midrule 
   DeepSeek-R1 First-Person & $0.793$ & $0.823$ & $\mathbf{0.841}$\\
   DeepSeek-R1 User & $-0.141$~~~ & $0.160$ & $\mathbf{0.445}$\\
   \midrule
   \midrule
   Llama 3.1-8B CoT First-Person       & $0.614$ & $0.784$ & $0.742$ \\
   Llama 3.1-8B CoT User               & $0.613$ & $0.806$ & $0.694$ \\
   Llama 3.1-8B Direct First-Person    & $0.611$ & $0.819$ & $0.682$ \\
   Llama 3.1-8B Direct User            & $0.593$ & $0.844$ & $0.686$ \\
   \midrule
   Llama 3.2-3B CoT First-Person       & $0.491$ & $0.604$ & $0.548$ \\
   Llama 3.2-3B CoT User               & $0.238$ & $0.549$ & $0.563$ \\
   Llama 3.2-3B Direct First-Person    & $0.293$ & $0.570$ & $0.599$ \\
   Llama 3.2-3B Direct User            & $0.372$ & $0.621$ & $0.490$ \\
   \midrule
   Gemma 3-4B CoT First-Person       & $0.292$ & $0.240$ & $0.513$ \\
   Gemma 3-4B CoT User               & $0.403$ & $0.572$ & $0.279$ \\
   Gemma 3-4B Direct First-Person    & $-0.017$~~~ & $0.104$ & $0.054$ \\
   Gemma 3-4B Direct User            & $0.472$ & $0.442$ & $0.219$ \\
   \bottomrule
 \end{tabular}
 }
\end{table}
Across all models and evaluation settings, alignment scores were highly consistent between Chain-of-Thought (CoT) and direct prompting, and between perspective prompting (first vs user) (Table~\ref{table:exp_2_corrs_full}). Statistical tests revealed no significant advantage for either approach ($p > .1$). Both prompting methods yield comparably stable correspondence with human judgments, suggesting that the observed patterns are not artifacts of prompting format, but rather reflect genuine model-level tendencies that remain robust across elicitation strategies.

\subsection{Analysis along dimensions of trust and incentives}
\label{app:exp2_analyses}
\begin{table}
\caption{Prompting strategy comparison across alignment dimensions. Alignment remains highly consistent across prompting styles, with no statistically significant differences between Chain-of-Thought and direct prompting, nor with respect to perspective (user or first-person).}
\label{table:exp2_prompting_alignment}
\centering
\begin{minipage}{0.48\textwidth}
\centering
\textbf{(a) Intra-character reward alignment}\vspace{0.5em}

\resizebox{0.9\textwidth}{!}{
\begin{tabular}{lr}
\toprule
Model & Correlation \\
\midrule
Llama First CoT & $0.996$ \\
DeepSeek-R1 First & $0.995$ \\
Gemini 2.0 Flash User Direct & $0.994$ \\
Claude User CoT & $0.994$ \\
Gemini 2.0 Flash First Direct & $0.994$ \\
Llama User CoT & $0.994$ \\
Claude First CoT & $0.994$ \\
GPT-4o First Direct & $0.993$ \\
o1 First & $0.993$ \\
GPT-4o First CoT & $0.993$ \\
Claude User Direct & $0.992$ \\
GPT-4o User CoT & $0.991$ \\
Claude First Direct & $0.991$ \\
Gemini 2.0 Flash User CoT & $0.992$ \\
Gemini 2.0 Flash First CoT & $0.991$ \\
GPT-4o User Direct & $0.987$ \\
Llama First Direct & $0.980$ \\
Llama User Direct & $0.970$ \\
o1 User & $0.947$ \\
o3-mini First & $0.932$ \\
o3-mini User & $0.851$ \\
DeepSeek-R1 User & $0.472$ \\
\bottomrule
\end{tabular}
}
\end{minipage}%
\hfill
\begin{minipage}{0.48\textwidth}
\centering
\textbf{(b) Cross-character calibration}\vspace{0.5em}

\resizebox{0.9\textwidth}{!}{
\begin{tabular}{lr}
\toprule
Model & Correlation \\
\midrule
Claude First Direct & $0.966$ \\
GPT-4o User Direct & $0.948$ \\
Claude User Direct & $0.948$ \\
GPT-4o User CoT & $0.945$ \\
GPT-4o First Direct & $0.944$ \\
Gemini 2.0 Flash User Direct & $0.941$ \\
Llama First Direct & $0.937$ \\
GPT-4o First CoT & $0.936$ \\
Claude First CoT & $0.931$ \\
Llama First CoT & $0.926$ \\
Gemini 2.0 Flash First Direct & $0.923$ \\
Gemini 2.0 Flash First CoT & $0.918$ \\
Gemini 2.0 Flash User CoT & $0.918$ \\
Claude User CoT & $0.918$ \\
Llama User Direct & $0.917$ \\
Llama User CoT & $0.909$ \\
o1 First & $0.909$ \\
DeepSeek-R1 First & $0.841$ \\
o1 User & $0.813$ \\
o3-mini First & $0.772$ \\
o3-mini User & $0.667$ \\
DeepSeek-R1 User & $0.445$ \\
\bottomrule
\end{tabular}
}
\end{minipage}
\end{table}

To investigate deeper into how each prompting factor --- reasoning (direct/CoT) and perspective --- affects the ability of LLMs to exercise motivational vigilance, we compute intra-character alignment as the average of the individual character-wise correlations between final reward-quality judgments, and cross-character calibration as the global correlation between final quality predictions across all characters and incentives. Results are in Table~\ref{table:exp2_prompting_alignment}. 

We find that frontier LLMs and reasoning models are typically more rational with respect to differences in incentive amounts (e.g., \$10 vs.~ \$1000) than differences in in characters (e.g., online source vs.~friend). Correlations with respect to differences in incentives ranged from $0.996$ to $0.472$, whereas correlations with respect to differences in trustworthiness of characters ranged from $0.969$ to $0.445$. Importantly, all three reasoning models tested performed among the worst in correlations both within and across characters. This suggests that these reasoning models may be ill-prepared to take on agentic tasks where the environment can contain ill-motivated communication (e.g., online advertisements).
\newpage
\section{Supplementary Details: Experiment 3}

\subsection{Prompts}
\label{app:exp3_prompts}

\subsubsection{GPT-4o censor prompt}
Prompt: 
\begin{quote}
    You will be given a sponsor message from a video transcript. Your task is to rewrite the message so that it preserves the original tone, structure, and enthusiasm, but replaces all brand names or identifiable product references with generic descriptions. The format and style should remain virtually identical.

    If the transcript does not describe a sponsor message (i.e., there is no actual sponsorship content), return the exact string: NO\_SPONSOR

    Return only the modified transcript or NO\_SPONSOR. DO NOT return any explanations, commentary, or extra text.

    Examples:

    Original:
    "This video is sponsored by Raid Shadow Legends – one of the biggest mobile RPGs of all time!"
    Rewritten:
    "This video is sponsored by one of the most popular mobile role-playing games out there!"
    
    Original:
    "Thanks to NordVPN for sponsoring today's video. Get secure and private access to the internet with NordVPN!"
    Rewritten:
    "Thanks to a leading virtual private network service for sponsoring today's video. Get secure and private access to the internet with this VPN!"
    
    Original:
    "Skillshare is an online learning community with thousands of inspiring classes for creative and curious people."
    Rewritten:
    "This video is brought to you by an online learning platform with thousands of inspiring classes for creative and curious people."
    
    Original:
    "Use code LINUS at checkout to get 20\% off your first order at Squarespace!"
    Rewritten:
    "Use the creator’s promo code at checkout to get 20\% off your first order at a popular website-building platform!"
    
    Original:
    "This episode is sponsored by BetterHelp — get matched with a licensed therapist and start living a happier life today."
    Rewritten:
    "This episode is sponsored by an online therapy service — get matched with a licensed professional and start living a happier life today."
    
    Original:
    "Smash that like button and don't forget to subscribe!"
    Rewritten:
    NO\_SPONSOR

    Now rewrite the following sponsor message:
    
    \{TRANSCRIPT\}
\end{quote}

\subsubsection{Product quality prompts}

First-Person: \begin{quote}You are watching a YouTube video called \{video\_title\}. The video was posted by the YouTuber \{channel\_name\}. Their channel's description is as follows: \{channel\_description\}. Their video contains the sponsor message in the following transcript: "\{sponsor\_message\_transcript\}"
On a scale from 0 to 100, with 0 meaning \{offer\_ratings[0]\} and 100 meaning \{offer\_ratings[1]\}, what do you think about the quality of the product promoted in the video?\end{quote}

User: \begin{quote}I am watching a YouTube video called \{video\_title\}. The video was posted by the YouTuber \{channel\_name\}. Their channel's description is as follows: \{channel\_description\}. Their video contains the sponsor message in the following transcript: "\{sponsor\_message\_transcript\}"
On a scale from 0 to 100, with 0 meaning \{offer\_ratings[0]\} and 100 meaning \{offer\_ratings[1]\}, what do you think about the quality of the product promoted in the video?\end{quote}

offer\_ratings = ["product is much worse than alternatives", "product is much better than alternatives"]

Direct prompting add-on: \begin{quote}Please only respond with an integer score, do not include anything else.\end{quote}

CoT prompting add-on: \begin{quote}Please reason about your answer before providing it. Structure your response in the exact following format: 

    Reasoning: [...]
    
    Answer (integer score only): [...]\end{quote}

\subsubsection{Reward quality prompts}

First-Person: \begin{quote} You are watching a YouTube video called \{video\_title\}. The video was posted by the YouTuber \{channel\_name\}. Their channel's description is as follows: \{channel\_description\}. Their video contains the sponsor message in the following transcript: "\{sponsor\_message\_transcript\}"
On a scale from 0 to 100, rate how beneficial you think this sponsor deal was for the YouTuber, with 0 meaning \{reward\_ratings[0]\} and 100 meaning \{reward\_ratings[1]\}, based on how the YouTuber would likely perceive its value. 
\end{quote}

User: \begin{quote}I am watching a YouTube video called \{video\_title\}. The video was posted by the YouTuber \{channel\_name\}. Their channel's description is as follows: \{channel\_description\}. Their video contains the sponsor message in the following transcript: "\{sponsor\_message\_transcript\}"
On a scale from 0 to 100, rate how beneficial you think this sponsor deal was for the YouTuber, with 0 meaning \{reward\_ratings[0]\} and 100 meaning \{reward\_ratings[1]\}, based on how the YouTuber would likely perceive its value.\end{quote}

reward\_ratings = ["this sponsor deal would not matter at all", "this sponsor deal would be extremely good to get"]

Direct prompting add-on: \begin{quote}Please only respond with an integer score, do not include anything else.\end{quote}

CoT prompting add-on: \begin{quote}Please reason about your answer before providing it. Structure your response in the exact following format: 

    Reasoning: [...]
    
    Answer (integer score only): [...]\end{quote}

\subsubsection{Trustworthiness prompts}

First-Person: \begin{quote}
    You are watching a YouTube video called \{video\_title\}. The video was posted by the YouTuber \{channel\_name\}. Their channel's description is as follows: \{channel\_description\}. Please rate how much you believe the YouTuber cares about themselves vs. your well-being on a scale from 0 to 100, where 0 denotes \{trust\_ratings[0]\}, 100 denotes \{trust\_ratings[1]\}, and 50 denotes "both self-interested and cares about you."
\end{quote}

trust\_ratings = ["entirely self interested", "only cares about you"]

User: \begin{quote}
    I am watching a YouTube video called \{video\_title\}. The video was posted by the YouTuber \{channel\_name\}. Their channel's description is as follows: \{channel\_description\}. Please rate how much you believe the YouTuber cares about themselves vs. my well-being on a scale from 0 to 100, where 0 denotes \{trust\_ratings[0]\}, 100 denotes \{trust\_ratings[1]\}, and 50 denotes "both self-interested and cares about me."
\end{quote}

trust\_ratings = ["entirely self interested", "only cares about me"]

Direct prompting add-on: \begin{quote}Please only respond with an integer score, do not include anything else.\end{quote}

CoT prompting add-on: \begin{quote}Please reason about your answer before providing it. Structure your response in the exact following format: 

    Reasoning: [...]
    
    Answer (integer score only): [...]\end{quote}

\subsection{Additional Steering Interventions}

We devise two new steering prompts that highlight other relevant approaches to vigilance: Gricean and Bias-oriented. Due to cost limitations, all experiments are conducted with GPT-4o. The new steering prompts are as follows:

Gricean: \texttt{When answering, consider what the speaker is trying to achieve by recommending this product. What are their likely goals or interests in this context?}

Bias-Oriented: \texttt{Before forming your answer, evaluate whether the recommendation might be biased. What motivations or incentives could be shaping the speaker’s advice?}

The Gricean prompt, based on Grice's \citep{grice1975logic} work on communicative maxims, cues potential self-interest in speech to foster vigilance. CoT is more effective at internal rationalization than direct prompting, though the improvements—linked to goal-oriented reasoning—are smaller than with our original incentives-based steering prompt.

For the bias-oriented steering prompt, we observe more consistent increases in correlation across all prompting methods, although the magnitudes of the correlation increases are still noticeably lower than what we obtained with the original incentives-based steering prompt.

\begin{table}
  \caption{Correlations between influence scores of LLMs with different steering prompts and the Bayesian model fitted to those LLMs' elicited priors.}
  \label{table:youtube_steering_additional}
  \centering
  \resizebox{\textwidth}{!}{
  \begin{tabular}{lSSSS}
    \toprule
    {} & \multicolumn{4}{c}{Correlation with Bayesian inference model} \\
    \cmidrule(r){2-5}
    Prompting Combination & {Default Prompt} & {Incentives Steer} & {Gricean Steer} & {Bias Steer} \\
    \midrule
    CoT First-Person         &  0.0240  &  \textbf{0.1367} & 0.0849  & 0.0759 \\
    CoT User                 &  0.0082  &  \textbf{0.1431} & 0.1793  & 0.1007 \\
    Direct First-Person     &  0.1211  &  \textbf{0.2338} & -0.0184 & 0.0901 \\
    Direct User             & -0.0056  &  \textbf{0.3121} & -0.0097 & 0.2321 \\
    \bottomrule
  \end{tabular}
  }
\end{table}

Together, these new results support that while the original steering prompt was relatively simple, it is especially effective at activating motive-sensitive reasoning in LLMs, outperforming alternative framings that target similar conceptual constructs.


\end{document}